\PassOptionsToPackage{unicode}{hyperref}
\PassOptionsToPackage{hyphens}{url}
\PassOptionsToPackage{dvipsnames,svgnames,x11names}{xcolor}

\documentclass[12pt]{article}
\usepackage{amsmath,amssymb}
\usepackage{array}
\usepackage{booktabs}
\usepackage{graphicx}
\usepackage{enumerate}
\usepackage{mathrsfs}
\usepackage{mathtools}
\usepackage{multicol}
\usepackage{multirow}
\usepackage{comment}
\usepackage{booktabs,threeparttable}
\usepackage{bm}
\usepackage{placeins}
\newcommand{\cond}{\mathrm{cond}}
\usepackage{amsfonts}

\usepackage[linesnumbered,ruled]{algorithm2e}
\newtheorem{theorem}{Theorem}

\newtheorem{lemma}{Lemma}
\newtheorem{assumption}{Assumption}

\usepackage{natbib}
\usepackage{url} 
\usepackage[pagewise]{lineno}

\usepackage{iftex}
\ifPDFTeX
  \usepackage[T1]{fontenc}
  \usepackage[utf8]{inputenc}
  \usepackage{textcomp} 
\else 
  \usepackage{unicode-math}
  \defaultfontfeatures{Scale=MatchLowercase}
  \defaultfontfeatures[\rmfamily]{Ligatures=TeX,Scale=1}
\fi
\usepackage{lmodern}
\ifPDFTeX\else  
\fi
\IfFileExists{upquote.sty}{\usepackage{upquote}}{}
\IfFileExists{microtype.sty}{
  \usepackage[]{microtype}
  \UseMicrotypeSet[protrusion]{basicmath} 
}{}
\makeatletter
\@ifundefined{KOMAClassName}{
  \IfFileExists{parskip.sty}{%
    \usepackage{parskip}
  }{
    \setlength{\parindent}{0pt}
    \setlength{\parskip}{6pt plus 2pt minus 1pt}}
}{
  \KOMAoptions{parskip=half}}
\makeatother
\usepackage{xcolor}
\makeatletter
\ifx\paragraph\undefined\else
  \let\oldparagraph\paragraph
  \renewcommand{\paragraph}{
    \@ifstar
      \xxxParagraphStar
      \xxxParagraphNoStar
  }
  \newcommand{\xxxParagraphStar}[1]{\oldparagraph*{#1}\mbox{}}
  \newcommand{\xxxParagraphNoStar}[1]{\oldparagraph{#1}\mbox{}}
\fi
\ifx\subparagraph\undefined\else
  \let\oldsubparagraph\subparagraph
  \renewcommand{\subparagraph}{
    \@ifstar
      \xxxSubParagraphStar
      \xxxSubParagraphNoStar
  }
  \newcommand{\xxxSubParagraphStar}[1]{\oldsubparagraph*{#1}\mbox{}}
  \newcommand{\xxxSubParagraphNoStar}[1]{\oldsubparagraph{#1}\mbox{}}
\fi
\makeatother

\usepackage{longtable,booktabs,array}
\usepackage{calc} 
\usepackage{etoolbox}
\makeatletter
\patchcmd\longtable{\par}{\if@noskipsec\mbox{}\fi\par}{}{}
\makeatother
\IfFileExists{footnotehyper.sty}{\usepackage{footnotehyper}}{\usepackage{footnote}}
\makesavenoteenv{longtable}
\usepackage{graphicx}
\makeatletter
\def\maxwidth{\ifdim\Gin@nat@width>\linewidth\linewidth\else\Gin@nat@width\fi}
\def\maxheight{\ifdim\Gin@nat@height>\textheight\textheight\else\Gin@nat@height\fi}
\makeatother
\setkeys{Gin}{width=\maxwidth,height=\maxheight,keepaspectratio}
\makeatletter
\def\fps@figure{htbp}
\makeatother

\makeatletter
\@ifpackageloaded{caption}{}{\usepackage{caption}}
\AtBeginDocument{%
\ifdefined\contentsname
  \renewcommand*\contentsname{Table of contents}
\else
  \newcommand\contentsname{Table of contents}
\fi
\ifdefined\listfigurename
  \renewcommand*\listfigurename{List of Figures}
\else
  \newcommand\listfigurename{List of Figures}
\fi
\ifdefined\listtablename
  \renewcommand*\listtablename{List of Tables}
\else
  \newcommand\listtablename{List of Tables}
\fi
\ifdefined\figurename
  \renewcommand*\figurename{Figure}
\else
  \newcommand\figurename{Figure}
\fi
\ifdefined\tablename
  \renewcommand*\tablename{Table}
\else
  \newcommand\tablename{Table}
\fi
}
\@ifpackageloaded{float}{}{\usepackage{float}}
\floatstyle{ruled}
\@ifundefined{c@chapter}{\newfloat{codelisting}{h}{lop}}{\newfloat{codelisting}{h}{lop}[chapter]}
\floatname{codelisting}{Listing}

\makeatother
\makeatletter
\@ifpackageloaded{caption}{}{\usepackage{caption}}
\@ifpackageloaded{subcaption}{}{\usepackage{subcaption}}
\makeatother

\ifLuaTeX
  \usepackage{selnolig}  
\fi
\usepackage[]{natbib}
\usepackage{bookmark}

\IfFileExists{xurl.sty}{\usepackage{xurl}}{} 
\hypersetup{
  pdftitle={A Bayesian Generative Modeling Approach for Arbitrary Conditional Inference},
  pdfauthor={Qiao Liu; Wing Hung Wong},
  pdfkeywords={Deep latent variable model; Bayesian deep learning; Stochastic optimization; Data imputation; Markov chain Monte Carlo},
  colorlinks=true,
  linkcolor={blue},
  filecolor={Maroon},
  citecolor={Blue},
  urlcolor={Blue},
  pdfcreator={LaTeX via pandoc}}

\newcommand{\anon}{1}

\begin{document}

\def\spacingset#1{\renewcommand{\baselinestretch}%
{#1}\small\normalsize} \spacingset{1}


\if1\anon
{
  \title{\bf An AI-powered Bayesian Generative Modeling Approach for Arbitrary Conditional Inference}
  \author{Qiao Liu\hspace{.2cm}\\
    Department of Biostatistics, Yale University\\
    and \\
    Wing Hung Wong \\
    Department of Statistics, Stanford University}
  \maketitle
} \fi

\if0\anon
{
  \bigskip
  \bigskip
  \bigskip
  \begin{center}
    {\LARGE\bf An AI-powered Bayesian Generative Modeling Approach for Arbitrary Conditional Inference}
\end{center}
  \medskip
} \fi

\bigskip
\begin{abstract}
Modern data analysis increasingly requires flexible conditional inference $P(X_{\mathcal{B}}\mid X_{\mathcal{A}})$, where $(X_{\mathcal{A}},X_{\mathcal{B}})$ is an arbitrary partition of the observed variables $X$. Existing approaches are either restricted to a fixed conditioning structure or depend strongly on the distribution of conditioning masks during training. To address these limitations, we introduce Bayesian generative modeling (BGM), a unified framework for arbitrary conditional inference. BGM learns a generative model of $X$ via a stochastic iterative Bayesian updating algorithm in which model parameters and latent variables are updated until convergence. Once trained, any conditional distribution can be obtained without retraining. Empirically, BGM achieves superior predictive performance with posterior predictive intervals, demonstrating that a single learned model can serve as a universal engine for conditional prediction with principled uncertainty quantification. We provide theoretical guarantees for convergence of the stochastic iterative algorithm, statistical consistency, and conditional risk bounds. The proposed BGM framework leverages modern AI to capture complex relationships among variables while adhering to Bayesian principles, offering a promising approach for a wide range of applications in modern data science. Code for BGM is available at \url{https://github.com/liuq-lab/bayesgm}. Document of BGM is available at \url{https://bayesgm.readthedocs.io}.
\end{abstract}

\noindent%
\vfill

\newpage
\spacingset{1.8} 

\section{Introduction}
\label{sec:intro}

Conditional inference concerns the problem of inferring the distribution of a response variable given a set of predictor variables, a fundamental task in both machine learning and statistical science~\citep{reid1995roles}. Classical supervised learning typically focuses on predicting a response variable given a fixed set of predictors. However, many modern data analysis tasks require arbitrary conditional inference, where one seeks to estimate $P(X_{\mathcal{B}}\mid X_{\mathcal{A}})$ for any partition $(X_{\mathcal{A}},X_{\mathcal{B}})$ of the observed variables $X$. Such flexibility is essential in dynamic real-world scenarios with heterogeneous or time-varying observation pattern. Discriminative conditional prediction methods typically lack this flexibility as they cannot perform new conditional inference without modifying the model architecture or retraining from scratch when the conditioning set changes.

To enable arbitrary conditional inference, early classical approaches utilized linear latent variable models and finite mixtures, such as the Mixture of Factor Analyzers (MFA) \citep{ghahramani1996algorithm}, which models $P(X)$ as a finite mixture of low-rank Gaussian components and thereby supports arbitrary conditional distributions in closed-form. To relax the assumption of a fixed number of mixture components, nonparametric Bayesian models such as Dirichlet process mixtures (DPMs) \citep{ferguson1973bayesian,antoniak1974mixtures} place priors on infinite mixtures. While DPMs offer a more flexible framework for
joint density estimation and arbitrary conditioning, the posterior computation remains challenging and scaling to large, high-dimensional datasets is nontrivial.

To better capture nonlinear relationships, non-linear latent variable models such as the Gaussian Process Latent Variable Model (GPLVM) \citep{lawrence2005probabilistic} and its Bayesian extensions (Bayesian GPLVM) \citep{titsias2010bayesian} emerged as powerful alternatives. By placing a Gaussian Process prior on the mapping from a low-dimensional latent space to the observed data space, GPLVMs provide a flexible, nonparametric approach to modeling complex joint distributions. These methods face severe computational bottlenecks due to exact Gaussian Process inference, making it difficult to scale and apply to modern large-scale data where both the sample size and feature dimension are substantial.

Recent years have witnessed rapid progress in conditional inference driven by advances in generative artificial intelligence (AI). This has led to the development of AI models to handle varying conditionings, such as Arbitrary Conditional Evaluation (ACE) \citep{strauss2021arbitrary}, Variational Autoencoders with Arbitrary Conditioning (VAEAC) \citep{ivanov2018variational}, and Arbitrary Conditional Flows (ACFlow) \citep{li2020acflow}. By training neural networks across random masking patterns or employing conditional normalizing flows, these models can capture highly nonlinear relationships and approximate arbitrary conditional prediction. However, these AI-driven approaches either heavily rely on the distribution of masking during training process or impose architectural constraints to neural networks. Furthermore, these methods focus primarily on accurate density estimation or point prediction. They often lack a coherent statistical mechanism for principled uncertainty quantification, which is critical in high-stakes domains where decisions must account for variability.

To address this widespread lack of rigorous uncertainty quantification in predictive modeling, a separate line of research has focused on conformal prediction (CP), which provides a model-agnostic framework for constructing prediction sets with guaranteed finite-sample coverage~\citep{vovk2005algorithmic,lei2018distribution,barber2021predictive}. Conformal prediction methods transform point predictors into valid predictive intervals or regions without relying on parametric assumptions, ensuring that the true response lies within the constructed set with a user-specified probability. Recent developments have substantially expanded the scope of conformal prediction methods, including quantile-based approaches~\citep{romano2019conformalized}, as well as localized conformal prediction~\citep{guan2023localized}, which adapts coverage to heterogeneous data distributions. These CP approaches have become a cornerstone for uncertainty quantification in modern machine learning applications due to their simplicity and strong theoretical guarantees~\citep{kato2023review}. However, while CP methods address the need for uncertainty quantification, they remain fundamentally constrained by the fixed conditioning structure. Furthermore, conformal prediction typically offers marginal coverage rather than full conditional calibration.

To address above challenges, we propose a Bayesian Generative Modeling (BGM) framework that leverages the power of modern AI techniques while adhering to the foundational principles of Bayesian inference. BGM learns the underlying generative process of the observed variables from a low-dimensional latent space, estimated by a stochastic iterative updating algorithm. Once trained, BGM enables arbitrary conditional inference, predicting the distribution of any subset of variables given the remaining without retraining or modifying the model architecture. This “train once, infer anywhere’’ property allows BGM to naturally accommodate dynamically changing conditioning sets that commonly arise in practice. By adopting a coherent Bayesian principle, BGM delivers posterior predictive intervals at any user-specified significance level. The proposed BGM framework thus combines the flexibility of modern AI models with the statistical coherence of Bayesian inference, offering a powerful and scalable solution for conditional prediction and uncertainty quantification in high-dimensional settings. We summarize the contributions of BGM as follows.

\begin{itemize}
    \item BGM formulates arbitrary conditional inference as posterior updating in a AI-powered Bayesian latent variable model, providing a unified probabilistic framework that fundamentally extends existing conditional inference methods, which are typically limited to one fixed conditioning structure or require a training mask distribution.

    \item We develop a stochastic iterative updating algorithm with convergence guarantee and establish posterior consistency and conditional risk bounds under mild regularity conditions. It also ensures efficient scaling to large, high-dimensional datasets since only a mini-batch of data is required at each step during training and posterior inference for each test sample is independent.

    \item We demonstrate that expressive neural parameterizations can be embedded within this Bayesian framework. Extensive experiments demonstrate that BGM achieves superior predictive accuracy with uncertainty quantification compared to leading conformal prediction methods and data imputation methods. These results highlight BGM’s practical value as a principled tool for conditional inference in modern data analysis.
    
\end{itemize}

\section{Methods}
\label{sec:meth}
\subsection{Problem Setup}

Considering an observational study with \textit{i.i.d.} observations of $\{X^{i}|i=1,...,N\}$ drawn from an unknown distribution $P(X)$ where $X\in \mathbb{R}^p$ denotes a $p$-dimensional random vector of observed variables. We are interested in estimating the conditional distribution of a subset of variables $X_{\mathcal{B}}$ given the remaining subset $X_{\mathcal{A}}$ where $(X_{\mathcal{A}}, X_{\mathcal{B}})$ represents an arbitrary partition of $X$. In a typical regression setting, $X_{\mathcal{A}}\in \mathbb{R}^{p_1}$ corresponds to the predictor variables, $X_{\mathcal{B}}\in \mathbb{R}^{p_2}$ is known as the response variables, and we have $p_1+p_2=p$. $X_{\mathcal{A}}$ is typically multivariate and $X_{\mathcal{B}}$ can be either a scalar or multivariate variable.

More generally, we define a random partition of variable $X=(X_1,...,X_p)$ as two disjoint subsets $X_{\mathcal{A}}$, $X_{\mathcal{B}}$ such that the associated index sets $\mathcal{A}$, $\mathcal{B}$ satisfy (1) $\mathcal{A}, \mathcal{B} \neq \emptyset$, (2) $\mathcal{A}, \mathcal{B} \subseteq \{1, \ldots, p\}$, (3) $\mathcal{A} \cap \mathcal{B} = \emptyset$, and (4) $\mathcal{A} \cup \mathcal{B} = \{1, \ldots, p\}$. This formulation unifies a broad class of statistical problems, including classical regression, data reconstruction, missing data imputation where the partition may vary across tasks, allowing arbitrary conditional inference over subsets of variables.

Although estimating the expected response $\mathbb{E}[X_{\mathcal{B}}|X_{\mathcal{A}}]$ remains central to most regression analysis, this statistic alone provides an incomplete picture of the underlying uncertainty. Practitioners often require much richer distributional information, such as conditional variances, quantiles, tail probabilities, or predictive intervals, to assess risk, reliability, and variability in predictions. Therefore, our focus extends beyond point estimation to the broader goal of accurately characterizing the full conditional distribution of interest, providing both point estimate and uncertainty quantification across arbitrary subsets of variables to condition on.

\subsection{Generative Process of Observed Data}

\begin{figure}
\begin{center}
\includegraphics[width=4.in]{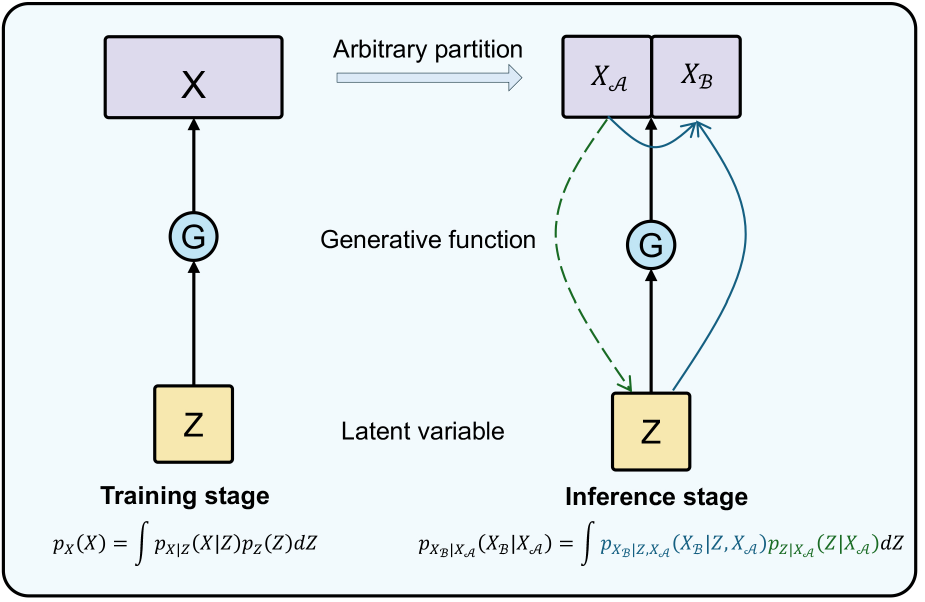}
\end{center}
\caption{The overview of BGM model. In the training stage, BGM serves as a generative model to learn the distribution of $P_X(X)$. Once fitted, BGM is capable of conditional inference under arbitrary partition of the observed variable $X$ in the inference stage. Variables are in rectangles and functions are in circles with incoming arrows indicating inputs to the function and outgoing arrows indicating outputs. The dotted arrow (green) represents Bayesian inference for the posterior of latent variable $Z$. \label{fig:model}}
\end{figure}

The BGM model is described in Figure~\ref{fig:model}, where $X$ represents observed variables and $Z$ denotes the low-dimensional latent variable that needs to be inferred. The generative process of observed data is formulated as 
\begin{equation}
\label{eqn:generative_model}
\left\{
\begin{aligned}
Z \sim& \pi_Z(Z),\\
\theta\sim& \pi_{\theta}(\theta),\\
X \sim& P(X|Z;\theta),\\
\end{aligned}
\right.
\end{equation}
where $\theta$ is a common parameter to specify all the conditional distribution of $X$ given $Z$. The prior distributions of both $Z$ and $\theta$ are set to be multivariate normal distributions. By default, we model the conditional distribution of $X|Z$ as normal distributions for continuous variables and logistic regression for discrete variables. In a typical continuous setting, the generative processes are defined as follows
\begin{equation}
\label{eqn:gen_cov}
P(X|Z;\theta) = \mathcal{N}(\mu(Z),\Sigma(Z)),
\end{equation}
where both mean and covariance matrix are learnable functions of latent variable $Z$ parameterized by $\theta$. In practice, we can further simplify the covariance matrix $\Sigma(Z)$ as a diagonal structure $\Sigma(Z)=diag(\sigma^2_1(Z),...,\sigma^2_p(Z))$. In this diagonal simplification, $X_{\mathcal{A}}$ is independent of $X_{\mathcal{B}}$ given the latent variable $Z$. A richer covariance structures can capture residual conditional dependence (see our Discussion). Note that the generative function $G$ provides both mean function $\mu(Z)$ and covariance function $\Sigma(Z)$, which is represented by a neural network with two output heads parametrized by $\theta$. To account for uncertainty or variation in model parameter space, we can also adopt Bayesian neural network (BNN), where model parameters are treated as random variables instead of deterministic values~\citep{goan2020bayesian,jospin2022hands}.

Note that we do not directly assume that the observed high-dimensional $X$ follow a multivariate Gaussian distribution in the data generative process, . Instead, our model learns the conditional distribution of $X$ given latent variable $Z$. This distinction is important: the generative model assumes a Gaussian form conditionally, not marginally. Importantly, the distributional assumption (e.g., Gaussian) primarily affects the modeling of the conditional variance, not the mean. Therefore, even if the true data-generating process deviates from a multivariate Gaussian distribution, the learned mean function shall remain robust.

\subsection{Stochastic Iterative Updating Algorithm} 
\label{sec:meth_iterative}

We designed a stochastic iterative algorithm to update the model parameters $\theta$ and the latent variable $Z$ until convergence. According to Bayes' theorem, the joint posterior distribution of the latent variable $Z$ and model parameters $\theta$ is represented as
\begin{equation}\label{eqn:posterior}
\begin{aligned}
&P(Z,\theta|X)=P(\theta|X)P(Z|X,\theta). \\
\end{aligned}
\end{equation}
Since the true posterior joint distribution is not tractable, we approximate the target by the following iterative procedure. Specifically, we iteratively (i) update the latent variable $Z$ from $P(Z|X, \theta)$. (ii) update model parameters $\theta$ from $P(\theta|X,Z)$. 

To update latent variable $Z$ in step (i), we denote the log-posterior of the latent variable $Z$ as 
\begin{equation}\label{eqn:posterior_Z}
\begin{aligned}
logP(Z|X,\theta)=log\pi_Z(Z)+logP(X|Z;\theta)+C, \\
\end{aligned}
\end{equation}
where $C=log\pi_{\theta}(\theta)-logP(X,\theta)$ does not involve $Z$. The second term in the log-posterior (\ref{eqn:posterior_Z}) represents the log-likelihood of the generative model. Under diagonal covariance structure, it is further denoted as 
\begin{equation}
\label{eqn:ll_models}
\begin{aligned}
logP(X|Z;\theta)=-\frac{1}{2}\sum_{i=1}^p[log(\sigma^2_i(Z))+(X_i-\mu_i(Z))^2/\sigma^2_i(Z)]+C_1,\\
\end{aligned}
\end{equation}
where $C_1=C-\frac{p}{2}log(2\pi)$, $X_i$ is the $i$-th element in $X$, and $\mu_i(Z)$ is the $i$-th element in the mean vector $\mu(Z)$. We update latent variable $Z$ for each sample during the training stage by maximizing (\ref{eqn:posterior_Z}) through stochastic gradient ascent. Note that the log-posterior of the latent variable $Z$ in (\ref{eqn:posterior_Z}) can be fully decoupled as the summation of per-sample contribution as 
\begin{equation}
\label{eqn:per_sample}
\begin{aligned}
logP(X|Z;\theta)=\sum_{i=1}^{N_{train}}logP(X^i|Z^i;\theta),\\
\end{aligned}
\end{equation}
where $N_{train}$ is the training sample size. The update of $Z$ can be done for each data point independently of other data points.

To update model parameters $\theta$ of the generative model in step (ii), the log posterior of $\theta$ can be represented as 
\begin{equation}
\label{eqn:posterior_theta}
\begin{aligned}
logP(\theta|X,Z)=log\pi_{\theta}(\theta)+logP(X|Z;\theta)+C_2,\\
\end{aligned}
\end{equation}
where $C_2$ is constant in $\theta$. To account for variation in model parameters instead of treating them as deterministic values, we can also employ a Bayesian neural network, which uses variational inference (VI)~\citep{blei2017variational} to approximate (\ref{eqn:posterior_theta}). Specifically, a variational distribution $q_{\phi}(\theta)$ is introduced to approximate the true posteriors in (\ref{eqn:posterior_theta}) where $q_{\phi}(\theta) \sim \mathcal{N}(\theta|\mu_{\phi}, \sigma_{\phi}^2I_p)$. Note that $\phi=(\mu_{\phi},\sigma_{\phi}^2)$ are learnable parameters for the variational distribution. The evidence lower bound (ELBO) for the posterior is given by 
\begin{equation}
\label{eqn:elbo}
\begin{aligned} 
\mathcal{L}(\phi)=\mathbb{E}_{q_{\phi}(\theta)}[logP(X|Z;\theta)]-KL(q_{\phi}(\theta)||\pi_{\theta}(\theta)),\\
\end{aligned}
\end{equation}
where the first term represents the expected log-likelihood under the variational posterior, and the second term is the Kullback–Leibler divergence between the variational distribution and the prior over model parameters. To enable efficient gradient-based optimization with respect to the variational parameters ($\phi$), we adopt the reparameterization trick~\citep{kingma2015variational}, which is denoted as
\begin{equation}
\label{eqn:flipout}
\begin{aligned}
\hat{\theta}=\mu_{\phi}+\sigma_{\phi}\odot\epsilon,\\
\end{aligned}
\end{equation}
where $\epsilon \sim \mathcal{N}(0,I_{d})$. $\odot$ is the element-wise product. $d$ is the dimensionality of the parameter space of the generative model. Applying variational inference in BNNs using mini-batches can result in high-variance gradient estimates. To mitigate this issue, we adopt the Flipout technique \citep{wen2018flipout} for implementation of the reparameterization trick. Flipout reduces gradient variance by decorrelating model parameters perturbations across different training examples within the same mini-batch. Specifically, instead of using a single shared random perturbation for all examples, Flipout generates pseudo-independent perturbations for each data point in the mini-batch, thereby decorrelating the resulting gradients, reducing their variance, and improving training stability. 

In each iteration, we first update the variational parameters by maximizing the ELBO in (\ref{eqn:elbo}), and then draw samples of the model parameters according to (\ref{eqn:flipout}). Conditioned on these sampled parameters, we can use a standard forward pass through the network, which includes linear transformations followed by nonlinear activation functions, to compute the mean and variance functions $\mu(Z)$ and $\Sigma(Z)$ or their derivatives. Importantly, each iteration requires only a randomly sampled mini-batch of observed data. Within each iteration, we first compute the gradient of (\ref{eqn:posterior_Z}) with respect to the latent variable $Z$ and update $Z$ for each individual using stochastic gradient ascent where an Adam~\citep{kingma2014adam}) optimizer is used conditional on the current model parameters. We then compute the gradient of each ELBO term in (\ref{eqn:elbo}) with respect to the variational parameters $\phi$ of the generative model, and update $\phi$ via stochastic gradient ascent to maximize the ELBO given the current latent variables.

\subsection{Arbitrary Conditional Inference}

Once the BGM model is fitted through the above iterative algorithm. We can use a trained BGM model for arbitrary conditional inference in the inference stage without retraining or modifying the model architecture. Let $(X_{\mathcal{A}}, X_{\mathcal{B}})$ represent an arbitrary partition of $X$, the conditional distribution can be denoted as 
\begin{equation}
\label{eqn:cond_dis_1}
\begin{aligned}
P_{X_{\mathcal{B}} \mid X_{\mathcal{A}}}(X_{\mathcal{B}} \mid X_{\mathcal{A}})
= \int P_{X_{\mathcal{B}} \mid Z, X_{\mathcal{A}}}(X_{\mathcal{B}} \mid Z, X_{\mathcal{A}})
\, P_{Z \mid X_{\mathcal{A}}}(Z \mid X_{\mathcal{A}})\, dZ,\\
\end{aligned}
\end{equation}
we tackle the above integral (\ref{eqn:cond_dis_1}) through two steps. In the first step, we draw multiple latent variable $Z$ from the posterior distribution $p_{Z \mid X_{\mathcal{A}}}(Z \mid X_{\mathcal{A}})$ using Markov chain Monte Carlo (MCMC) reviewed in \cite{geyer2011introduction}. In the second step, we draw $X_{\mathcal{B}}$ from $p_{X_{\mathcal{B}} \mid Z, X_{\mathcal{A}}}(X_{\mathcal{B}} \mid Z, X_{\mathcal{A}})$ based on all Monte Carlo samples of $Z$ where $p_{X_{\mathcal{B}} \mid Z, X_{\mathcal{A}}}(X_{\mathcal{B}} \mid Z, X_{\mathcal{A}})$ has a closed form as a multivariate Gaussian distribution. 

In the first step of sampling the posterior samples of latent variable $Z$, we adopt Hamiltonian Monte Carlo (HMC) algorithm, which uses gradient-informed dynamical trajectories to make long-distance proposals with high acceptance rates, achieving faster and better sampling efficiency~\citep{neal2011mcmc}. According to Bayes Theorem, the log posterior of latent variable $Z$ can be represented as 
\begin{equation}
\label{eqn:Z_posterior}
\begin{aligned}
logP_{Z \mid X_{\mathcal{A}}}(Z \mid X_{\mathcal{A}})=log\pi_Z(Z)+logP_{X_{\mathcal{A}}|Z}(X_{\mathcal{A}}|Z)+C_3\\
\end{aligned}
\end{equation}
where $C_3=-logP_{X_{\mathcal{A}}}(X_{\mathcal{A}})$ and $X_{\mathcal{A}}|Z\sim \mathcal{N}(\mu_{\mathcal{A}}(Z),\Sigma_{\mathcal{A}}(Z))$. Here, $\mu_{\mathcal{A}}(Z)$ and $\Sigma_{\mathcal{A}}(Z))$ are the corresponding partition components in the $\mu(Z)$ and $\Sigma(Z)$ for $X_{\mathcal{A}}$, respectively. Under the diagonal covariance simplification, $\Sigma_{\mathcal{A}}(Z)$ also has a diagonal structure. For each test data point for $X_{\mathcal{A}}$, we keep a sufficient number of MCMC posterior samples of latent variable $Z$. Note that sampling low-dimensional latent variable $Z$ for each test data point is fully decoupled, which enables efficient parallelization and further improving computational efficiency. We implemented MCMC sampling process fully on GPU in parallel by using TensorFlow Probability (TFP) library~\citep{dillon2017tensorflow} for acceleration. 

In the second step, we need to sample $X_{\mathcal{B}}$ given $X_{\mathcal{A}}$ and latent variable $Z$, which is available in closed-form because of the Gaussianity of joint distribution of $X_{\mathcal{A}}$ and $X_{\mathcal{B}}$. Specifically, since
\begin{equation}
\label{eqn:decompose}
\begin{aligned}
\begin{pmatrix}
X_{\mathcal A} \\
X_{\mathcal B}
\end{pmatrix}
\sim
\mathcal N\!\left(
\begin{pmatrix}
\mu_{\mathcal A} \\
\mu_{\mathcal B}
\end{pmatrix},
\begin{pmatrix}
\Sigma_{\mathcal A\mathcal A} & \Sigma_{\mathcal A\mathcal B} \\
\Sigma_{\mathcal B\mathcal A} & \Sigma_{\mathcal B\mathcal B}
\end{pmatrix}
\right),\\
\end{aligned}
\end{equation}
we have
\begin{equation}
\label{eqn:cond_dis}
\begin{aligned}
X_{\mathcal{B}} \mid Z, X_{\mathcal{A}}\sim \mathcal{N}(\mu_{\mathcal{B}|\mathcal{A}}(Z),\Sigma_{\mathcal{B}|\mathcal{A}}(Z)),\\
\end{aligned}
\end{equation}
where
\begin{equation}
\label{eqn:conditional_mvg}
\left\{
\begin{aligned}
\mu_{\mathcal{B}|\mathcal{A}}(Z)= &\mu_{\mathcal B}(Z)
+ \Sigma_{\mathcal B\mathcal A}(Z)\,
  \Sigma_{\mathcal A\mathcal A}(Z)^{-1}\,
  \bigl(X_{\mathcal A} - \mu_{\mathcal A}(Z)\bigr),\\
\Sigma_{\mathcal{B}|\mathcal{A}}(Z)= &\Sigma_{\mathcal B\mathcal B}(Z)
- \Sigma_{\mathcal B\mathcal A}(Z)\,
  \Sigma_{\mathcal A\mathcal A}(Z)^{-1}\,
  \Sigma_{\mathcal A\mathcal B}(Z).\\
\end{aligned}
\right.
\end{equation}
Under the diagonal covariance simplification, the formula (\ref{eqn:conditional_mvg}) becomes
\begin{equation}
\label{eqn:conditional_mvg_sim}
\left\{
\begin{aligned}
\mu_{\mathcal{B}|\mathcal{A}}(Z)= &\mu_{\mathcal B}(Z),\\
\Sigma_{\mathcal{B}|\mathcal{A}}(Z)= &\Sigma_{\mathcal B\mathcal B}(Z).\\
\end{aligned}
\right.
\end{equation}

Through the above two-step sampling process, we can perform conditional inference by sampling from $P_{X_{\mathcal{B}} \mid X_{\mathcal{A}}}(X_{\mathcal{B}}|X_{\mathcal{A}})$. Let us denote the test data in inference stage as $\{X_{\mathcal{A}}^{i}|i=1,...,N_{test}\}$ where $N_{test}$ is the sample size in test set. We obtain the posterior samples of latent variable through the first step of sampling process, which are denoted as $\{Z^{i,m}|i=1,...,N_{test},m=1,...,N_{MC}\}$ where $N_{MC}$ is the MCMC sample size. In the second step of sampling, all the posterior samples of latent variable $Z$ are fed to the generative model $G$ to obtain the mean and covariance functions in (\ref{eqn:conditional_mvg_sim}) in order to sample from (\ref{eqn:cond_dis}). The final conditional posterior samples are denoted as $\{X_{\mathcal{B}}^{i,m}|i=1,...,N_{test},m=1,...,N_{MC}\}$. The point estimate for the prediction can be constructed based on the conditional mean of the posterior samples as
\begin{equation}
\label{eqn:posterior_mean}
\begin{aligned}
\hat{X}_{\mathcal{B}}^{i}=\frac{1}{N_{MC}}\sum_{m=1}^{N_{MC}}X_{\mathcal{B}}^{i,m}.\\
\end{aligned}
\end{equation}

The prediction interval can be constructed based on the conditional samples. Given a test data point $X_{\mathcal{A}}^{i}$ and desired significant level $\alpha$ (e.g., $\alpha=0.05$), we calculate the $\alpha/2$-quantile and $(1-\alpha/2)$-quantile to represent the lower and upper prediction interval bounds, respectively as follows
\begin{equation}
\label{eqn:quantile_ite}
\left\{
\begin{aligned}
\hat{L}_{\mathcal{B}}^i=&Quantile_{\alpha/2}(\{X_{\mathcal{B}}^{i,m}|m=1,...,N_{MC}\}),\\
\hat{U}_{\mathcal{B}}^i=&Quantile_{1-\alpha/2}(\{X_{\mathcal{B}}^{i,m}|m=1,...,N_{MC}\}).\\ 
\end{aligned}
\right.
\end{equation}
where $Quantile_{\alpha}(\cdot)$ is the quantile function of the sampling distribution that cuts off the lower $\alpha$ tail of the distribution.

\subsection{Model Initialization}

The parameters of neural net (e.g., weights and biases) are typically initialized by a uniform or normal distribution~\citep{narkhede2022review}. Inspired from our previous works \citep{Liu_pnas2024,liu2025ai}, the model performance can be further improved through an encoding generative modeling (EGM) initialization strategy. Specifically, an auxiliary pseudo-inverse encoder function $E$ is added to BGM to directly map the $X$ to the latent variable. Specifically, we desire that the distribution of $Z=E(V)$ should match the prior distribution of latent variable $Z$, which is a standard multivariate Gaussian distribution. The distribution match is achieved by adversarial training \citep{goodfellow2014generative,liu2021density}. By the encoding process, the high-dimensional covariates with unknown distribution are mapped to a low-dimensional latent space with a desired distribution. Both the latent variable $Z$ and model parameters in $G$ are initialized through the EGM process. After initialization, the encoder function $E$ is removed during model training process.

\subsection{Model Hyperparameters}
For vector-valued datasets, generative model $G$ is implemented as a fully connected neural network with three hidden layers, each containing 128 units. We use the leaky-ReLu activation function ($LeakyReLU(x)=max(0.2x,x)$) in each hidden layer. The latent space is set to 5 for observational data within 100 dimension and 10 otherwise. Two parallel output heads produce the conditional mean and diagonal variance where the variance head uses the Softplus activation ($Softplus(x)=log(1+e^x)$) to ensure strictly positive variance.

For MNIST image data~\citep{deng2012mnist}, The generative model $G$ adopts a fully convolutional decoder network that maps a 10-dimensional latent space to the $28\times 28$ image space. The latent vector $Z$ is first passed through a fully connected layer and reshaped into a low-resolution feature map of size $7\times7\times4F$, where $F$ is the number of base kernels (default $F=32$). The feature map is then progressively upsampled via two transposed convolution layers with stride 2, producing a $28\times28$ spatial resolution, followed by an additional convolution layer for refinement. Each convolution block is equipped with batch normalization~\citep{ioffe2015batch} and LeakyReLU activations. Two parallel $1\times1$ convolution output heads generate the pixel-wise mean and variance, with Softplus applied to the variance head to ensure positivity.

We train BGM using the Adam optimizer \citep{kingma2014adam} with learning rate $0.005$ for both latent-variable updates and model-parameter updates. Training proceeds in mini-batches of size 32 for a default total of 500 epochs. To obtain a well-behaved initialization of both the latent space and model parameters, and to prevent early variance inflation, we adopt an EGM initialization as a warm start. This initialization phase uses up to 50,000 randomly sampled mini-batches prior to running the alternating stochastic iterative updating algorithm described in Section~\ref{sec:meth_iterative}.

During inference, we draw posterior samples of the latent variable $Z$ using Hamiltonian Monte Carlo (HMC). The HMC sampler is initialized with step size 0.01 and an adaptive step-size procedure targeting an acceptance rate of approximately 0.75. For each test point, we discard the first 5,000 transitions as burn-in and retain 5,000 posterior samples for the latent variable. All HMC chains are fully vectorized and executed in parallel across observations, enabling efficient large-scale conditional inference.

\section{Theoretical Results}

\subsection{Convergence of the stochastic iterative updating algorithm}

We analyze the convergence of the stochastic iterative updating algorithm described in section \ref{sec:meth_iterative}. For a mini-batch of samples at iteration $t$ ($t\in[0,...,T-1]$), we first update latent variables of the current batch by stochastic ascent of the log posterior in (\ref{eqn:posterior_Z}), then we update the variational parameters $\phi=(\mu_{\phi},\sigma_{\phi}^2)$ by stochastic ascent of the ELBO term in (\ref{eqn:elbo}) using the reparameterization in (\ref{eqn:flipout}). The parameters $\theta$ for generative model G are sampled from $q_{\phi}$ between these two updates. 

Let $w=(Z,\phi)$ where $Z=(Z^1,..,Z^N)$ represents the latent variables of all samples. The objective function can be written as 
\begin{equation}
\label{eqn:obj_func}
\begin{aligned}
\mathcal{J}(w)
= \frac{1}{N}\sum_{i=1}^N \left\{
\mathbb{E}_{q_{\phi}(\theta)}[\log P(X^i \mid Z^i; \theta)]
+ \log \pi_Z(Z^i)
\right\}
- \mathrm{KL}\!\left(q_{\phi}(\theta)\,\|\,\pi_{\theta}(\theta)\right),\\
\end{aligned}
\end{equation}
where the iterative algorithm performs stochastic block coordinate ascent on $\mathcal{J}$. At iteration $t$, we use a mini-batch $B_t \subset \{1,\ldots,N\}$ of size $|B_t| = m$. The natural mini-batch surrogate is 
\begin{equation}
\begin{aligned}
J_{B_t}(w)
\;:=\;
\frac{1}{m}\sum_{i \in B_t} \ell_i(w)
\;-\;
\mathrm{KL}\!\left(q_\phi(\theta)\,\|\,\pi_\theta(\theta)\right),
\end{aligned}
\end{equation}
where $\ell_i(w)
\;:=\;
\mathbb{E}_{q_\phi(\theta)}
\!\left[
\log P(X^i \mid Z^i; \theta)
\right]
\;+\;
\log \pi_Z(Z^i)$ is the per-sample contribution for the objective function.

We further denote the joint state as $w_t=(Z_t, \phi_t)$ at iteration $t$, write the block‑stochastic gradients as $g_t^{(Z)}=\nabla_{Z} \mathcal{J}_{B_t}(w_t)$ and $g_t^{(\phi)}=\nabla_{\phi} \mathcal{J}_{B_t}(w_t)$, and denote the learning rates $\eta_t^{(Z)}$ and $\eta_t^{(\phi)}$.

Then the iteration at $t+1$ can be written compactly as
\begin{equation}
\label{eqn:ite_equ}
\left\{
\begin{aligned}
Z_{t+1} &= Z_{t}+\eta_t^{(Z)}g_t^{(Z)},\\
\phi_{t+1} &= \phi_{t}+\eta_t^{(\phi)}g_t^{(\phi)}.\\ 
\end{aligned}
\right.
\end{equation}

We introduce the main assumptions used in Stochastic approximation~\citep{robbins1951stochastic} and nonconvex stochastic optimization~\citep{ghadimi2013stochastic} as follows.
\begin{assumption}\label{as:smoothness}
The sequence $\{w_t\}$ remains almost surely in a compact set $\mathcal{W}\subset \mathbb{R}^{d_z + d_\phi}$. Denote full-data gradient of $\mathcal{J}$ by $\nabla \mathcal{J}=(\nabla_Z \mathcal{J},\nabla_{\phi} \mathcal{J})$ . $\mathcal{J}< \infty$ and $\mathcal{J}$ has $L$-Lipschitz gradient:
$$\|\nabla \mathcal{J}(w) - \nabla \mathcal{J}(w')\| \le L \|w - w'\|.$$
\end{assumption}

\begin{assumption}\label{as:gradient_bound}
The stochastic gradients satisfy unbiasedness: $\mathbb{E}\!\left[g_t^{(Z)} \mid \mathcal{F}_t\right]=\nabla_Z J(w_t)$ and $\mathbb{E}\!\left[g_t^{(\phi)} \mid \mathcal{F}_t\right]=\nabla_\phi J(w_t)$ where $\mathcal{F}_t$ is the filtration generated by the stochastic iterative updating algorithm up to iteration $t$. There exists constant $\sigma_Z^{2}$ and $\sigma_\phi^{2}$ such that $\mathbb{E}\,\|g_t^{(Z)}| \mathcal{F}_t\|^{2} \le \sigma_Z^{2}$ and $\mathbb{E}\,\|g_t^{(\phi)}|\mathcal{F}_t\|^{2} \le \sigma_\phi^{2}$ for all $t$. 
\end{assumption}

\begin{assumption}\label{as:lr}
Block‑wise learning rates $\eta_t^{(Z)}$ and $\eta_t^{(\phi)}$ satisfy $\sum_t \eta_t^{(Z)} = \infty, \sum_t \bigl(\eta_t^{(Z)}\bigr)^{2} < \infty$ and $\sum_t \eta_t^{(\phi)} = \infty, \sum_t \bigl(\eta_t^{(\phi)}\bigr)^{2} < \infty.$
\end{assumption}
Assumption \ref{as:smoothness} holds in the default architecture of generative model $G$ on compact sets where both Leaky-ReLU and Softplus activation functions are globally Lipschitz. Assumption \ref{as:lr} is a standard Robbins–Monro stepsize condition \citep{robbins1951stochastic} in stochastic approximation algorithms, which can be satisfied by an optimizer with designed decaying learning rate. Then we have the following theorems for convergence under above assumptions.
\begin{theorem} \label{thm: Estimation_Error}
(\textbf{Convergence to stationary points}) Every limit point $w_*$ of the sequence $\{w_t\}$ is almost surely first‑order stationary.  
$$\lim_{t \to \infty} \|\nabla \mathcal{J}(w_t)\| = 0 \quad a.s.$$
\end{theorem}
The detailed proof is given in Appendix A.

\begin{theorem} \label{thm: Finite-time}
(\textbf{Finite‑time rate}) There exists a sequence of random indices $R_T\in \{0,...,T-1\}$ such that
$$\mathbb{E}\,\|\nabla \mathcal{J}(w_{R_T})\|^{2}
\;\le\;
\frac{\mathcal{J}^* - \mathcal{J}(w_0)}
     {\sum_{t=0}^{T-1} \eta_t}
\;+\;
\,\frac{
L \sum_{t=0}^{T-1}
\big(
(\eta_t^{(Z)})^2 \sigma_Z^2
+ (\eta_t^{(\phi)})^2 \sigma_\phi^2
\big)
}{2
\sum_{t=0}^{T-1}\eta_t
} \, ,$$
\end{theorem}
where $\mathcal{J}^*=sup_w\mathcal{J}(w)$ and $\eta_t=\min\{\eta_t^Z,\eta_t^{\phi}\}$. The detailed proof is given in Appendix B.

In the case when the step sizes are small constant $\eta_t=\eta_t^Z=\eta_t^{\phi}=\eta$ and the distribution above is uniform on $\{0,...,T-1\}$, this simplifies to 
$$\mathbb{E}\!\left[\|\nabla \mathcal{J}(w_R)\|^{2}\right]
\;\le\;
\frac{\mathcal{J}^\star - \mathcal{J}(w_0)}{T\,\eta}
\;+\;
\frac{L}{2}\,\eta\,(\sigma_Z^{2} + \sigma_\phi^{2}) ,$$
which has the standard nonconvex SGD stationarity rate $O(1/T) + O(\eta)$.

\subsection{Statistical Consistency of the Learned Generative Model}

For any variational parameter $\phi$, which determines the variational posterior 
$q_{\phi}(\theta)$ over model parameters, the \emph{observable law} induced by $\phi$ is
\begin{equation}
    P_{\phi}(x)
    \;=\;
    \iint P(x \mid z; \theta)\,\pi_{Z}(z)\,q_{\phi}(\theta)\,dz\,d\theta,
\end{equation}
where $\pi_{Z}$ is the Gaussian prior on the latent variable $Z$.  
Distinct $\phi$ values may induce the same observable law $P_{\phi}$, 
which is why we focus on the distribution itself.

Let $P_{0}$ denote the true distribution of the data.  
Our goal is to show that the observable law generated by the fitted BGM, 
denoted $P_{\hat{\phi}_{N}}$, converges to a \emph{pseudo-true observable law} 
$P^{\star}$ as $N \to \infty$, where
\begin{equation}
    \Phi^{\star}
    \;:=\;
    \arg\max_{\phi \in \Phi} \widetilde{\mathcal{J}}(\phi),
    \qquad 
    P^{\star} \in \{P_{\phi} : \phi \in \Phi^{\star}\}.
\end{equation}

Here $\widetilde{\mathcal{J}}(\phi)$ is the \emph{population objective}
\begin{equation}
    \widetilde{\mathcal{J}}(\phi)
    \;=\;
    \mathbb{E}_{P_{0}}\!\left[\,m(X; \phi)\,\right]
    \;-\;
    \mathrm{KL}\!\left(q_{\phi}\,\|\,\pi_{\theta}\right),
\end{equation}
and $m(x;\phi)$ is the profiled complete-data criterion
\begin{equation}
    m(x; \phi)
    \;=\;
    \sup_{z \in \mathbb{R}^{d_{z}}}
    \biggl\{
        \mathbb{E}_{q_{\phi}(\theta)}
        \!\left[\log P(x \mid z; \theta)\right]
        + \log \pi_{Z}(z)
    \biggr\}.
\end{equation}

We assume that although $\Phi^{\star}$ may contain multiple parameter values 
due to the unidentifiability of latent variable $Z$, the observable law induced by them is unique; 
this law is denoted $P^{\star}$.  
Under well model specification, $P^{\star} = P_{0}$.

Because neural network parameter space is unbounded and highly non-convex, 
uniform statistical control must be restricted to expanding but compact subsets 
of $\Phi$. Following standard practice in modern M-estimation~\citep{shen1994convergence,de2021review}, we work with a 
\emph{sieve sequence}
\[
\Phi_{1} \subseteq \Phi_{2} \subseteq \cdots \subseteq \Phi,
\qquad 
\bigcup_{N \ge 1} \Phi_{N} \text{ is dense in } \Phi,
\]
where $\Phi_{N}$ restricts weight norms, spectral norms, and enforces a variance 
floor/ceiling.

At sample size $N$, the training procedure outputs a fitted variational parameter
\[
\hat{\phi}_{N} \in \Phi_{N}.
\]

The estimator need not be a global optimizer; instead, we track its 
\emph{algorithmic suboptimality}
\begin{equation}
    \delta^{\mathrm{alg}}_{N}
    \;=\;
    \sup_{\phi \in \Phi_{N}}
    \widetilde{\mathcal{J}}_{N}(\phi)
    \;-\;
    \widetilde{\mathcal{J}}_{N}(\hat{\phi}_{N}),
\end{equation}
where
\begin{equation}
    \widetilde{\mathcal{J}}_{N}(\phi)
    \;=\;
    \frac{1}{N}\sum_{i=1}^{N} m(X_{i}; \phi)
    \;-\;
    \mathrm{KL}\!\left(q_{\phi} \,\|\, \pi_{\theta}\right)
\end{equation}
is the empirical analogue of $\widetilde{\mathcal{J}}(\phi)$.

The term $\delta^{\mathrm{alg}}_{N}$ captures the fact that training is stochastic, 
non-convex, and only approximately optimizes $\widetilde{\mathcal{J}}_{N}$.

For consistency, we require that the population objective value decreases 
whenever the induced observable distribution moves away from $p^{\star}$.  
Formally, define the \emph{population separation margin}
\begin{equation}
    \Delta(\varepsilon)
    \;=\;
    \sup_{\phi \in \Phi} \widetilde{\mathcal{J}}(\phi)
    \;-\;
    \sup_{\phi :\, d(P_{\phi}, P^{\star}) \ge \varepsilon}
        \widetilde{\mathcal{J}}(\phi),
\end{equation}
where $d$ denotes the \emph{bounded--Lipschitz distance} on probability laws over 
$\mathbb{R}^{p}$:
\begin{equation}
    d(P,Q)
    \;=\;
    \sup_{\|f\|_{\infty} \le 1,\, \mathrm{Lip}(f) \le 1}
    \left|
        \int f\, dP
        \;-\;
        \int f\, dQ
    \right|.
\end{equation}

We first state the conditions required for establishing law-level consistency.

\begin{assumption}[Uniform LLN on the sieve]\label{ass:ULLN}
\begin{equation}
    \omega_{N}
    \;:=\;
    \sup_{\phi \in \Phi_{N}}
    \bigl|
        \widetilde{\mathcal{J}}_{N}(\phi)
        -
        \widetilde{\mathcal{J}}(\phi)
    \bigr|
    \;\xrightarrow{p}\; 0.
\end{equation}
\end{assumption}

\begin{assumption}[Algorithmic suboptimality]\label{ass:algo}
\begin{equation}
    \delta^{\mathrm{alg}}_{N}
    \;\xrightarrow{p}\; 0.
\end{equation}
\end{assumption}

\begin{assumption}[Sieve bias]\label{ass:sieve_bias}
\begin{equation}
    r_{N}
    \;:=\;
    \sup_{\phi \in \Phi}
        \widetilde{\mathcal{J}}(\phi)
    \;-\;
    \sup_{\phi \in \Phi_{N}}
        \widetilde{\mathcal{J}}(\phi)
    \;\longrightarrow\; 0.
\end{equation}
\end{assumption}

\begin{assumption}[Population separation]\label{ass:sep}
For all $\varepsilon > 0$,
\begin{equation}
    \Delta(\varepsilon)
    \;:=\;
    \sup_{\phi \in \Phi} \widetilde{\mathcal{J}}(\phi)
    \;-\;
    \sup_{\phi:\, d(P_{\phi},P^{\star}) \ge \varepsilon}
        \widetilde{\mathcal{J}}(\phi)
    \;>\; 0.
\end{equation}
\end{assumption}

\medskip

\begin{theorem}[Law-level Consistency]\label{thm:law_consistency}
Under Assumptions \ref{ass:ULLN}--\ref{ass:sep}, the observable law induced by the 
fitted BGM satisfies
\begin{equation}
    d\!\left(P_{\hat{\phi}_{N}},\, P^{\star}\right)
    \;\xrightarrow{p}\; 0.
\end{equation}
Under well specification, $P^{\star} = P_{0}$, and the estimator is statistically 
consistent for the true data generating distribution.
\end{theorem}
The detailed proof is given in Appendix C.

\subsection{Conditional‑Risk Bounds for Arbitrary Conditional Inference}

Define an arbitrary partition as $X = (X_{\mathcal{A}}, X_{\mathcal{B}}) \in \mathcal{X}_{\mathcal{A}} \times \mathcal{X}_{\mathcal{B}} \subseteq \mathbb{R}^{|\mathcal{A}|} \times \mathbb{R}^{|\mathcal{B}|}$. Given the learned observable law from BGM fitting $P_{\hat{\phi}_{N}}$ and the pseudo‑true observable law $P^\star$, their induced conditionals are denoted as $g_{P_{\hat{\phi}_N}}(x_{\mathcal{A}})=P_{\mathcal{B}\mid \mathcal{A};\, P_{\hat{\phi}_{N}}}\!\left(\,\cdot \mid x_{\mathcal{A}} \right)$ and $g_{P^{\star}}(x_{\mathcal{A}})=P_{\mathcal{B}\mid \mathcal{A};\, P^\star}\!\left(\,\cdot \mid x_{\mathcal{A}} \right)$, respectively. Note that $g$ is a measurable rule that maps $\mathcal{X_A}\rightarrow P_{\mathcal{B}}$ and $P_{\mathcal{B}}$ is the space of probability laws on $\mathcal{X_B}$.

One option is to define the loss function in the prediction stage as a kernel score (KS) function $\ell_{KS}:\mathcal{X_B}\times P_{\mathcal{B}}\rightarrow[0,U]$, which is denoted as 
$$\ell_{KS}(y,r)
:= k(y,y)
   - 2\, \mathbb{E}_{y' \sim r}\!\left[ k(y, y') \right]
   + \mathbb{E}_{y',\,y'' \sim r}\!\left[ k(y', y'') \right] \qquad (\forall y\in \mathcal{X_B},r\in P_{\mathcal{B}}),$$
where $k$ is the kernel function (e.g., RBF kernel), $y$ represents the observed value of the response variable and $r$ represents the predicted conditional distribution of $\mathcal{X_B}$ given an $x_{\mathcal{A}}$ (e.g., $P_{\mathcal{B}\mid \mathcal{A};\, \hat{\phi}_N}\!\left(\,\cdot \mid x_{\mathcal{A}} \right)$). The loss function is the squared Maximum Mean Discrepancy (MMD)~\citep{gretton2012kernel} between the Dirac distribution at $y$ and the predictive distribution $r$ using kernel $k$.

We evaluate conditional risk on a closed set
$\mathcal{X}_{\mathcal{A}}^{\circ} \subseteq \mathbb{R}^{|\mathcal{A}|}$ where conditioning is well-posed:
\[
\inf_{x_{\mathcal{A}} \in \mathcal{X}_{\mathcal{A}}^{\circ}}
P_{\mathcal{A};\,P^\star}(x_{\mathcal{A}}) \;\ge\; c_0 \;>\; 0.
\]

Then, the prediction risk on $\mathcal{X}_{\mathcal{A}}^{\circ}$ is denoted as:
\[
\mathcal{R}_{P^\star}^{\circ}(g)
\;:=\;
\mathbb{E}_{(X_{\mathcal{A}}, X_{\mathcal{B}})\sim P^\star,\circ}
\bigl[\, \ell_{KS}\!\left(g(X_{\mathcal{A}}),\, X_{\mathcal{B}}\right) \bigr],
\]
where $P^{\star,\circ}(dx_{\mathcal{A}}, dx_{\mathcal{B}})
\;:=\;
P^{\star,\circ}_{\mathcal{A}}(dx_{\mathcal{A}})
\, P_{\mathcal{B}\mid \mathcal{A};\,P^\star}(dx_{\mathcal{B}} \mid x_{\mathcal{A}}).
$ and $P^{\star, \circ}_{\mathcal{A}}(\,\cdot\,)
\;:=\;
P^\star_{\mathcal{A}}(\,\cdot \mid \mathcal{X}^{\circ}_{\mathcal{A}}\,)$. Note that $\; g_{P^\star} = \arg\min_{g} \, \mathcal{R}^{\circ}_{P^\star}(g)$ as the Bayes optimal or the ``pseudo-true'' minimizer if the model is misspecified, then we have the following theorem:

\begin{theorem}[Conditional Excess Risk Bound]\label{thm:cond_risk}
The conditional excess risk can be bounded by: 
\[
\mathcal{R}_{P^\star}^{\circ}(g_{P_{\hat{\phi}_N}})
\;-\;
\mathcal{R}_{P^\star}^{\circ}(\hat{g})
\;\le\;
L_\ell \, \varepsilon_N^{\mathrm{cond}},
\]
where $\varepsilon_N^{\mathrm{cond}}=\sup_{x_{\mathcal{A}} \in \mathcal{X}_{\mathcal{A}}^{\circ}}
d\!\left(
P_{\mathcal{B}\mid \mathcal{A};\,P_{\hat{\phi}_N}}(\,\cdot \mid x_{\mathcal{A}}),
P_{\mathcal{B}\mid \mathcal{A};\,P^\star}(\,\cdot \mid x_{\mathcal{A}})
\right)$.
\end{theorem}
Note that $\varepsilon_N^{\mathrm{cond}}$ represents the uniform conditional discrepancy on $\mathcal{X}_{\mathcal{A}}^{\circ}$ and $L_\ell$ is a Lipschitz constant for the loss function $\ell_{KS}$. Under the law-level consistency in Theorem~3 and standard stability of conditional distributions under bounded--Lipschitz convergence on closed sets where $P_{A;P^\star} \ge c_0 > 0$, $\varepsilon_N^{\cond} \xrightarrow{p} 0$. Therefore, the conditional excess risk of $g_{P_{\hat{\phi}_N}}$ relative to $\hat{g}$ vanishes asymptotically.
The detailed proof is given in Appendix D.

\section{Empirical Results}
To demonstrate the performance of BGM algorithm, we conducted a series of empirical experiments based on both simulation and real datasets. We benchmarked BGM against the leading conformal prediction methods for conditional inference tasks. In the task of data imputation, we employed the MNIST, the handwritten digits image dataset to demonstrate that the imputation power offered by BGM. We also compared BGM to the widely used imputation methods to show the superior performance of BGM in imputation tasks. 

\subsection{Model Evaluation}
For the conditional prediction tasks, we evaluate both the point estimate and the uncertainty interval estimate inferred from BGM and competing methods. For evaluating point estimation, we use mean squared error (MSE) Pearson correlation coefficient (PCC), Spearman correlation coefficient  (SCC) as evaluation metrics. For evaluating interval estimation, we compute the interval length given a specific significance level $\alpha$, PCC and SCC between the prediction interval length and the oracle interval length of testing set are calculated. Additionally, average interval length, and empirical coverage rate are used for model evaluation. Here, we also evaluate empirical coverage on test data (marginal). In the data imputation experiments, we reported the relative improvement of classification accuracy as the evaluation metrics.

\subsection{Baseline Methods}
For point estimation in conditional prediction tasks, we compare BGM method against Linear Regression, Random Forest \citep{breiman2001random}, XGboost \citep{chen2016xgboost}, VAEAC \citep{ivanov2018variational} and a neural network predictor. The neural network architecture follows the default configuration recommended in localized conformal prediction (LCP) \citep{guan2023localized} as the default predictor. Similarly to BGM, we used the conditional mean of multiple imputations from VAEAC as the point estimate where the number of imputations equals to the number of posterior sample size of BGM (default: 5000). 

For interval estimation in conditional prediction tasks, BGM is benchmarked against eight different conformal prediction (CP) methods \citep{lei2018distribution,romano2019conformalized,guan2023localized} and generative AI method VAEAC for conditional inference. The conformal prediction provides finite-sample, distribution-free marginal coverage guarantees without imposing parametric assumptions on the data generation process. The performance of CP methods depends critically on the choice of nonconformity score computed on a calibration set. Similar to BGM, we constructed prediction intervals of VAEAC by taking the $\alpha$/2- and (1-$\alpha$/2)-quantiles based on multiple imputed predictions.

For data imputation tasks, we compared BGM to widely used imputation approaches, including mean imputation and MICE~\citep{van2011mice}. We provide implementation details of baseline methods in Appendix E.

\subsection{Conditional Prediction}
We designed a simulation study to evaluate the performance of BGM in conditional prediction to capture nonlinear conditional structure and heteroscedasticity. The data are generated from a low-rank latent variable model, where both the predictors and response depend on a shared latent factor $Z$. To avoid notation conflict, we denote the predictors in the simulation by
$V$ and the response by $R$. The simulation data generation process is as follows.

\begin{equation}
\label{eq:simulation}
\left\{
\begin{aligned}
Z &\sim \mathcal{N}\!\left(0,\, I_{k}\right), \\
V \mid Z &\sim \mathcal{N}\!\left(0.2\, Z A^\top,\; 0.1^2 I_{d}\right), \\
R \mid Z &\sim \mathcal{N}\!\left(\sin(Z w),\; \left(0.1 + 0.5\, \text{sigmoid}(Z u)\right)^{2}\right),
\end{aligned}
\right.
\end{equation}
where $A\in \mathbb{R}^{d\times k}$ is a randomly generated loading matrix from a standard normal distribution that induces a low-rank structure in the predictor space, $w,u\in \mathbb{R}^k$ are coefficient vectors sampled from standard normal distribution controlling the nonlinear mean and heteroscedastic noise of the response, $\text{sigmoid}(x)=\frac{1}{1+e^{-x}}$ denotes the logistic function, and $I_d,I_k$ are the identify matrices. This construction leads to substantial variation in conditional variance across the feature space, creating a challenging for uncertainty quantification for $P(R|V)$.

We generate $N=20,000$ observations as $\{(V^i,R^i)|i=1,...,N\}$. In the BGM setting, we have the partition pattern as $X=(X_{\mathcal{A}}, X_{\mathcal{B}})$ where $X_{\mathcal{A}}=V$, $X_{\mathcal{B}}=R$ and $p_1=d$, $p_2=1$. We randomly split the data into 80\% training set and 20\% testing set. During training stage, we concatenated $V$ and $R$ from training set and fed the data to BGM for learning the joint distribution through the stochastic updating algorithm. In the testing stage, we inferred the posterior distribution of $P(R|V)$ on the held-out testing set and evaluated the model performance on both point estimate and interval estimate. For conformal prediction (CP) methods, 20\% of the training set was further retained for calibration purposes. A neural network with three fully connected layers is used for prediction in all CP methods as suggested by \citep{guan2023localized}. To evaluate the influence of the dimensionality of observation data $X$, we varied the dimension $p=d+1$ from 50, to 100, and 300. The significance level $\alpha$ is set to be 0.05.

\renewcommand\arraystretch{0.5}
\begin{table}[t]
\centering
\caption{Comparison of point estimation performance across different methods. Metrics include mean squared error (MSE), Pearson correlation (PCC), Spearman correlation (SCC) at different dimensions $p$.}
\label{tab:point_estimate_comparison}
\begin{threeparttable}
\small
\begin{tabular}{llcccccc}
\toprule
Dimension & Metric & Linear Regression & Random Forest  & XGBoost & VAEAC & LCP$^{\ast}$ & BGM \\
\midrule
\multirow{3}{*}{$p=50$}
  & MSE$\downarrow$ & 0.601 & 0.186 & 0.192 & 0.178 & 0.183 & \textbf{0.167} \\
  & PCC$\uparrow$ & 0.281 & 0.847 & 0.841 & 0.854& 0.848 & \textbf{0.864} \\
  & SCC$\uparrow$ & 0.414 & 0.846 & 0.841 & 0.856& 0.848 & \textbf{0.863} \\
\midrule
\multirow{3}{*}{$p=100$} 
  & MSE$\downarrow$ & 0.620 & 0.405 & 0.565 & 0.239& 0.217 & \textbf{0.193} \\
  & PCC$\uparrow$ & 0.040 & 0.656 & 0.304 & 0.786& 0.814 & \textbf{0.832} \\
  & SCC$\uparrow$ & 0.038 & 0.668 & 0.346 & 0.797& 0.825 & \textbf{0.847} \\
\midrule
\multirow{3}{*}{$p=300$} 
  & MSE$\downarrow$ & 0.631 & 0.260 & 0.352 & 0.594& 0.212 & \textbf{0.181} \\
  & PCC$\uparrow$ & 0.059 & 0.790 & 0.680 & 0.462& 0.817 & \textbf{0.846} \\
  & SCC$\uparrow$ & 0.063 & 0.794 & 0.701 & 0.535& 0.824 & \textbf{0.851} \\
\bottomrule
\end{tabular}
\begin{tablenotes}[para,flushleft]
\footnotesize
\textit{Note}: $^{\ast}$Point estimates for localized conformal prediction (LCP) are obtained using the neural network architecture implemented in the LCP codebase as the default predictor.
\end{tablenotes}
\end{threeparttable}
\end{table}

In the point estimation experiments, BGM consistently demonstrates the best point prediction performance among all competing methods across three different evaluation metrics (Table~\ref{tab:point_estimate_comparison}). Traditional linear regression performs substantially worse in all settings, especially as the dimension increases. VAEAC performs well when $p=50$ and performance decreases significantly when $p=300$. The neural network adopted from LCP is the most robust baseline compared to other methods. BGM further improves the best baseline by achieving a relative reduction in MSE ranging from 6.2\% to 14.6\% for different observation dimension. In addition, BGM achieves consistently higher correlation measures (both PCC and SCC) over the strongest baseline. These results highlight BGM’s superior ability to capture complex structures in high-dimensional conditional prediction tasks.

\begin{table}[t]
\centering
\caption{Comparison of interval estimation performance across BGM and baseline methods. 
Metrics include Pearson correlation (PCC), Spearman correlation (SCC), empirical marginal coverage, 
and average prediction interval length (ave.PI) at different dimensions $p$. The the nominal coverage level is set to $1-\alpha=0.95$ with $\alpha=0.05$. For reference, the average oracle interval lengths are 1.427 ($p=50$), 1.446 ($p=100$), and 1.400 ($p=300$).}
\label{tab:interval_estimate_comparison}
\scriptsize
\begin{tabular}{lcccccccccc}
\toprule
Dimension & Metric$^{\ast}$ & CP & LCP & LW-CP & LW-LCP & QR-CP & LWQR-CP & QR-LCP & LWQR-LCP & BGM \\
\midrule

\multirow{4}{*}{$p=50$}
& PCC$\uparrow$   &0.020  &0.878 &0.909  &0.897  &0.756  &0.770  &0.732  &0.755  &\textbf{0.937}  \\
& SCC$\uparrow$   &0.016 &0.887 &0.927 &0.917  &0.763  &0.778  &0.735  &0.763  &\textbf{0.987}  \\
& Coverage        &0.980  &0.981  &0.981  &0.980  &0.981  &0.981  &0.981  &0.981  &\textbf{0.944}  \\
& ave.PI$\downarrow$ &2.301 &2.147 &2.059 &2.027 &2.132 &2.123 &2.147 &2.136 &\textbf{1.450}  \\

\midrule
\multirow{4}{*}{$p=100$}
& PCC$\uparrow$   &-0.057  &0.812  &0.831 &0.837  &0.709  &0.718  &0.700  &0.703  &\textbf{0.874}  \\
& SCC$\uparrow$   &-0.111 &0.786  &0.844  &0.845  &0.706  &0.715  &0.698  &0.701  &\textbf{0.935}  \\
& Coverage        &0.979  &0.978  &0.978  &0.977  &0.985  &0.985  &0.984  &0.984  &\textbf{0.950}  \\
& ave.PI$\downarrow$ &2.603 &2.340 &2.210 &2.170 &2.437 &2.435 &2.425 &2.434 &\textbf{1.576}  \\


\midrule
\multirow{4}{*}{$p=300$}
& PCC$\uparrow$   &-0.013 &0.598  &0.628  &0.617  &0.500  &0.514  &0.495  &0.545  &\textbf{0.863}  \\
& SCC$\uparrow$   &-0.018 &0.601  &0.658  &0.652  &0.518  &0.545  &0.517  &0.563  &\textbf{0.941}  \\
& Coverage        &0.981  &0.983  &0.984  &0.985  &0.989  &0.990  &0.990  &0.991  &\textbf{0.966}  \\
& ave.PI$\downarrow$ &2.699 &2.514 &2.355 &2.374 &2.885 &2.882 &2.883 &2.877 &\textbf{1.694}  \\

\bottomrule
\end{tabular}
\begin{tablenotes}[para,flushleft]
\footnotesize
\textit{Note}: $^{\ast}$Coverage values closest to the nominal level 0.95 are highlighted in bold.
\end{tablenotes}
\end{table}

\begin{figure}[!htbp]
\centering
\includegraphics[width=4.7in]{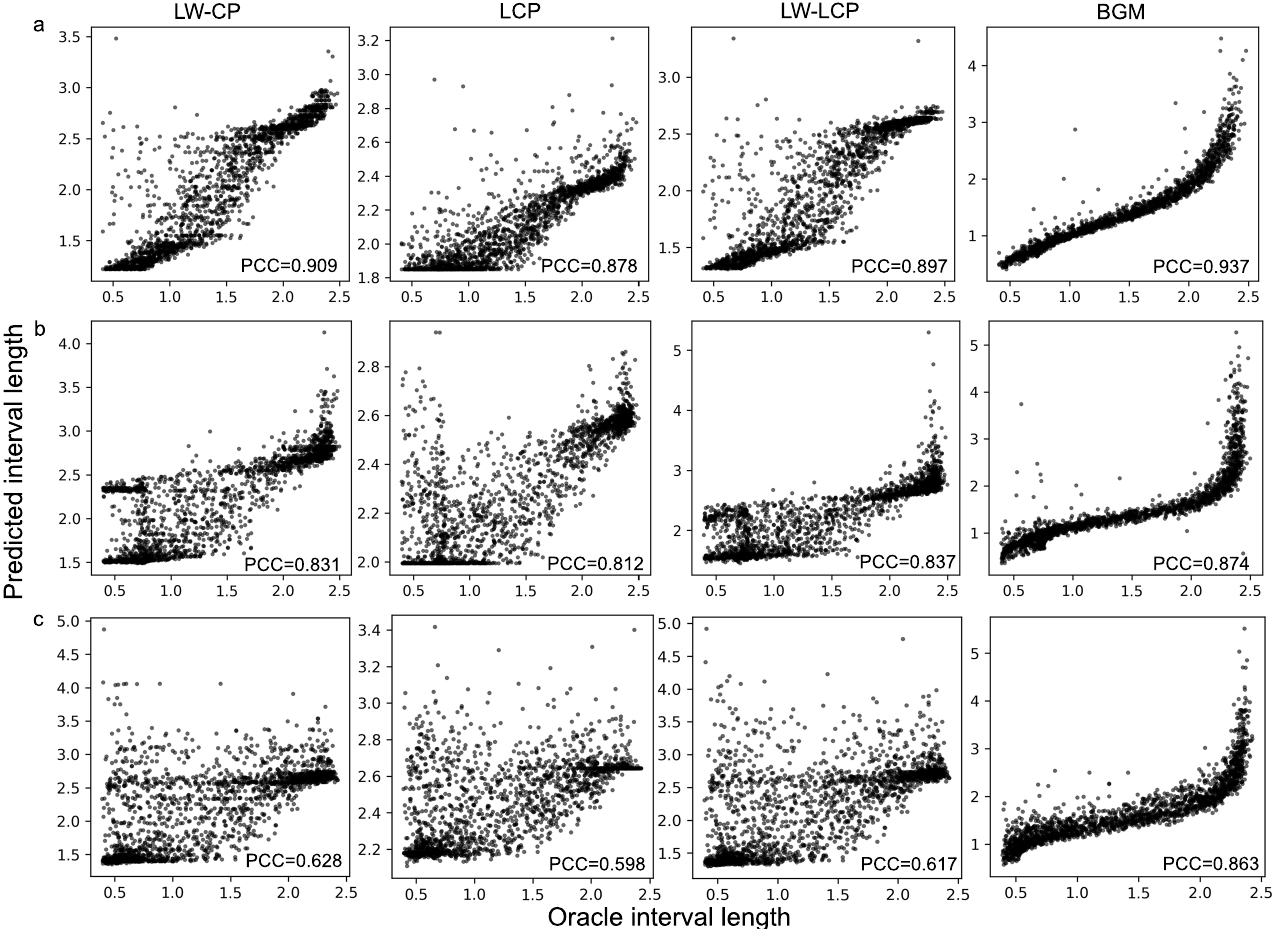}
\caption{The comparison of the estimated prediction intervals from BGM the top three conformal prediction methods. Each dot represents a point in the held-out testing set. (a) $p=50$; (b)$p=100$; (c)$p=300$.   \label{fig:PI_scatter}}
\end{figure}

Next, we evaluate the interval estimation to quantify the ability of capturing the uncertainty in the prediction tasks. As shown in Table~\ref{tab:interval_estimate_comparison}, BGM provides substantially better interval estimation than all conformal prediction (CP) baselines across all observational data dimensions. Vanilla CP fails to adapt to heteroscedastic noise because it relies on a single global residual quantile, resulting in nearly constant-width intervals across testing data points. Conformal prediction methods with localization largely improves the performance of standard CP by weighting calibration samples according a similarity-based localizer function relative to the test point, producing a local rather than global empirical distribution of nonconformity scores. Across all methods, BGM achieves the strongest alignment between the predicted and oracle interval lengths with the highest Pearson and Spearman correlations, ranging from 0.863 to 0.937 and 0.935 to 0.987, respectively, whereas the best CP baseline typically yields correlation between 0.6 and 0.9. In high dimensional setting (e.g., $p=300$), BGM substantially improves the PCC over the best CP baseline by 0.251, demonstrating its superior ability to adapt to underlying heteroscedasticity while distance-based localization of CP methods fails. In addition, BGM attains empirical coverage rate between 0.944 and 0.966, close to the nominal 95\% level, while all CP baselines are systematically more conservative with coverage from 0.980 to 0.991 but with considerably wider prediction intervals. We also benchmarked the performance of VAEAC in this task to demonstrate its inability to capture uncertainty (Table~\ref{tab:vaeac}).

To further examine the quality of the estimated prediction intervals, we compare BGM with the top three CP baselines by plotting the predicted interval lengths against the oracle interval lengths for all test points (Figure~\ref{fig:PI_scatter}). Here, oracle prediction interval is obtained via Monte Carlo integration under the true data-generating process. Specifically, for each test point ($v^i$), we first sample $Z^{i}\sim p(Z\mid V=v^i)$ using its closed-form Gaussian posterior. We draw $M$ Monte Carlo samples $\{Z^{i,m}\mid m=1,..,M\}$ of the latent variables, followed by sampling $R^{i,m}\sim p(R\mid Z=z^{i,m})$. The oracle interval is defined by the empirical $\alpha/2$- and (1-$\alpha/2$) quantiles of $\{R^{i,m}\mid m=1,..,M\}$. These scatter plots provide a visual assessment of calibration across different methods. Across all settings, BGM exhibits the strongest alignment with the oracle intervals, yielding substantially higher Pearson correlations than the competing methods, particularly in higher-dimensional settings. 

Overall, these results demonstrate that BGM not only provides accurate point estimates in the conditional prediction tasks but also achieves more accurate interval estimates than existing CP approaches with closer-to-oracle calibration of uncertainty. Together, this highlights BGM’s ability to deliver reliable and efficient predictions with uncertainty quantification, even in challenging high-dimensional and heteroscedastic settings.

\subsection{Data Imputation}

\begin{figure}[!htbp]
\begin{center}
\includegraphics[width=4in]{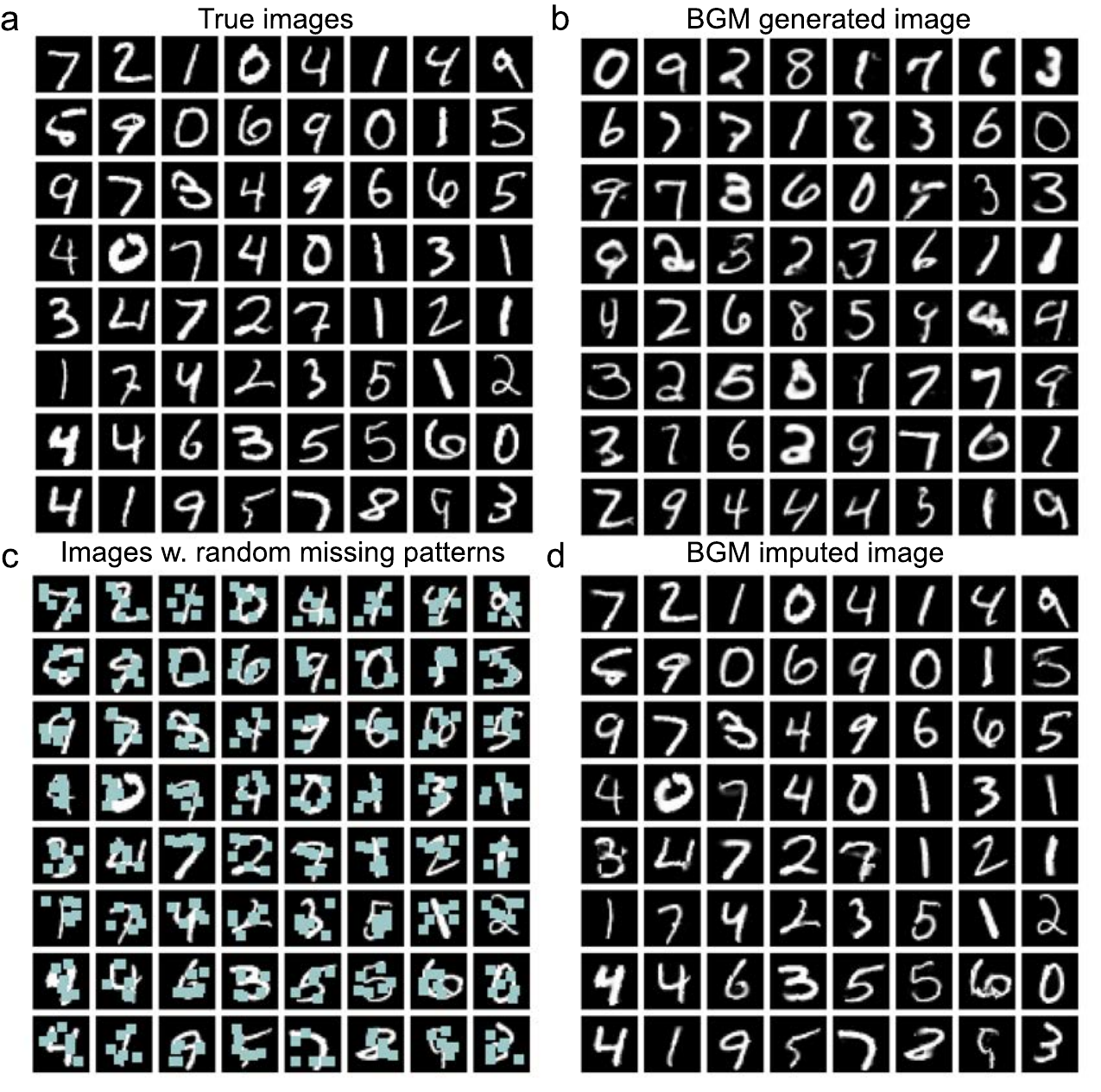}
\end{center}
\caption{Data imputation experiments on MNIST dataset with BGM. (a) True images from held-out testing set. (b) Generated images from BGM after model training. (c) Testing images with random missing patterns (six random $5\times5$ squares as masks). (d) The imputed results by a trained BGM model given (c) as input. The posterior mean is used for estimating each missing pixel. \label{fig:img_impute}}
\end{figure}

We further investigate the ability of BGM to impute missing values using the MNIST handwritten digits dataset. Each image is represented as a $28\times 28$ grayscale intensity vector, rescaled to $[0,1]$ (Figure~\ref{fig:img_impute}a). We train an BGM model on the MNIST training set with $60,000$ images, where the generative function $G$ is represented by a convolutional decoder network. After training, BGM learns the joint distribution of all pixels. We generated MNIST images with BGM by first randomly sampling from the prior distribution of the latent space, which is a standard Gaussian distribution. Then we obtained the mean and variance through the learned $G$ function. The generated MNIST images using the learned mean and variance functions well assemble the distribution of the true images (Figure~\ref{fig:img_impute}b).

Next, we used a single trained BGM model to impute missing pixels with arbitrary patterns through computing the conditional distribution of missing pixels given the observed ones. To create missingness, we start from held-out test images and randomly place six $5\times 5$ square masks on each image. All pixels in the masked regions are treated as missing (Figure~\ref{fig:img_impute}c). The BGM imputations were obtained by conditioning on the observed pixels and replacing each missing pixel by its posterior mean. Although the missing rate is as high as nearly 20\%, BGM reconstructs coherent digit shapes that retain both global digit identity and local stroke continuity (Figure~\ref{fig:img_impute}d and Figure~\ref{fig:Missingness}-\ref{fig:More_imp}).

\begin{figure}[!htbp]
\begin{center}
\includegraphics[width=5.0in]{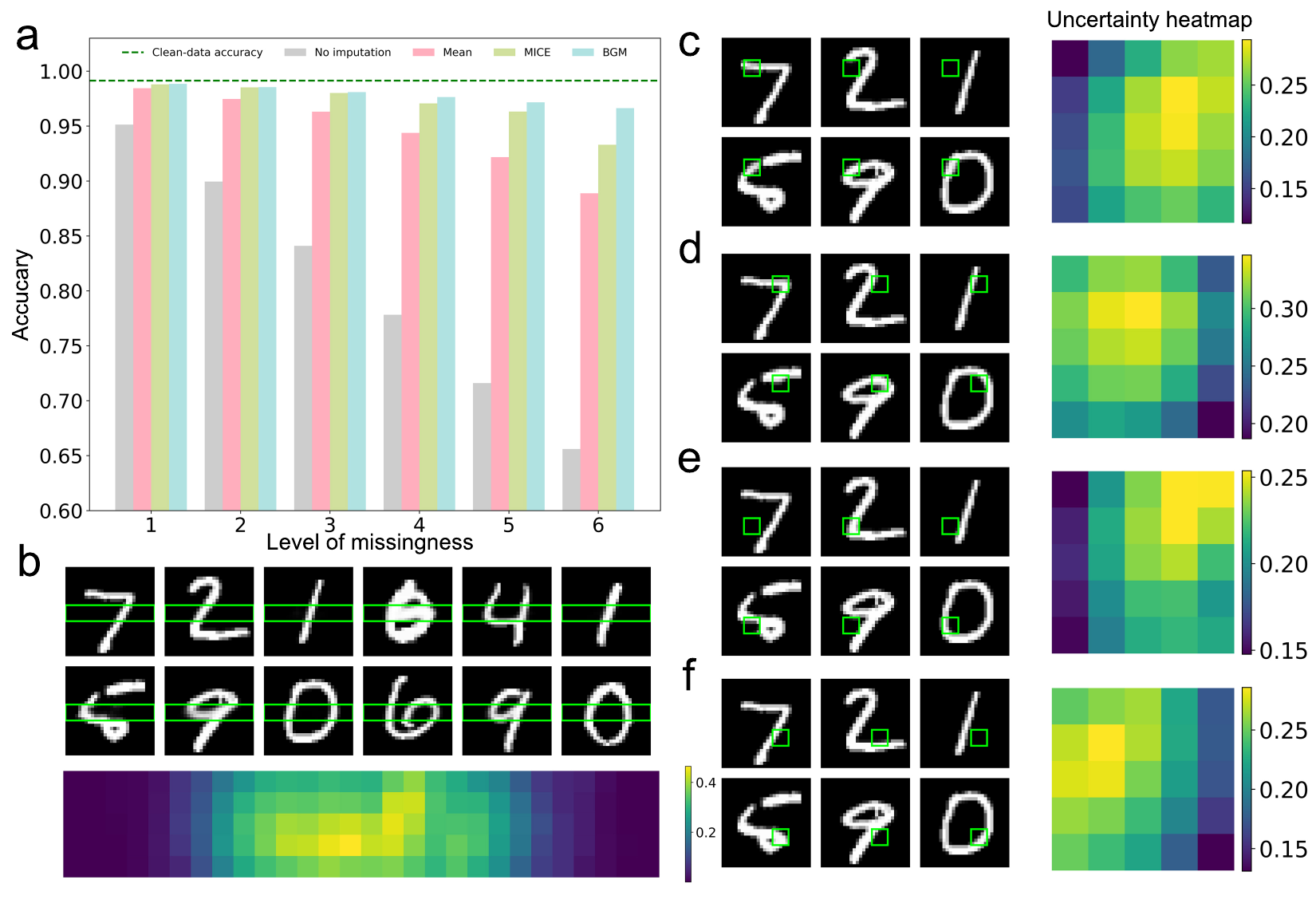}
\end{center}
\caption{Data imputation on MNIST dataset. (a) Imputation improves MNIST test classification accuracy by different methods with different level of missingness (e.g., number of random $5\times5$ holes). (b-f) BGM imputed images and uncertainty heatmap with different missing patterns. (b) A $5 \times 13$ stripe mask in the middle. (c) A $5 \times 5$ square mask in the upper left. (d) A $5 \times 5$ square mask in the upper right (e) A $5 \times 5$ square mask in the lower left. (f) A $5 \times 5$ square mask in the lower right. Note that the pixels within the green rectangles are imputed by BGM posterior mean and the uncertainty is denoted by the average prediction interval length with $\alpha=0.05$ across all test images. \label{fig:img_uncertainty}}
\end{figure}

To quantitatively assess the utility of BGM imputations for downstream prediction, we trained a standard convolutional neural network (CNN) classifier based on the original, fully observed MNIST training set. On clean test images this classifier achieves an accuracy of $0.9914$. We then evaluate the classification performance under six levels of increasing missingness on the testing set. Classification accuracy deteriorates rapidly as more pixels are removed, dropping from $0.9513$ to $0.6560$ as missingness increases. We next compared several imputation strategies applied prior to classification. As shown in Figure~\ref{fig:img_uncertainty}a, BGM imputations consistently provide the most substantial accuracy improvement across all masking levels, achieving accuracies of $0.966$ to $0.988$ and outperforming the classical imputation baselines.

Furthermore, we checked whether the uncertainty provided by BGM can offer additional information. We showcased different missing patterns in the testing images and visualized the uncertainty for the missed pixels. The uncertainty of imputed images demonstrated some interesting patterns (Figure~\ref{fig:img_uncertainty}b-f). In general, pixels closer to the image boundary have relatively smaller uncertainty, which is consistent with the fact that near-boundary pixels are more likely to be the ``black'' background. 

These results demonstrate that BGM learns a realistic conditional distribution over observed image pixels and can act as a powerful plug-in data imputer. Moreover, BGM offers the full posterior distribution for missing data imputations, providing much richer information and more flexibility for the missing patterns compared to traditional data imputers.


\section{Conclusion}
\label{sec:conc}
In this work, we introduce Bayesian Generative Models (BGM) as a highly flexible and powerful framework for conditional inference that leverages the power of AI while adhering to the Bayesian principles. By iteratively updating the model parameters for both mean and covariance functions and the low-dimensional latent variables, BGM learns the generative process of the observational data through latent variable modeling. Once the BGM model is trained, it could be applied to conditional prediction tasks with arbitrary partition for the observational data without retraining or modifying model architecture. 

We also established theoretical guarantees for the BGM framework, including consistency and asymptotic risk control under mild regularity conditions, showing that generative modeling can be competitive with methods that are tailored directly to regression. Empirically, across a range of simulated and real-data experiments, BGM consistently delivered accurate point predictions and uncertainty quantification, outperforming strong discriminative and conformal prediction baselines in various settings. The data imputation experiments on MNIST dataset further illustrated that a single trained BGM can serve as a versatile engine for data imputation under arbitrary missing patterns, benefiting downstream tasks, such as classification.

There are several directions for future improvement of BGM. First, our model offers the posterior distribution in the tasks of conditional prediction. How to fully utilize the distributional information to benefit downstream statistical or machine learning tasks requires further investigation. Second, more complex covariance structures, such as low-rank setting, can be incorporated into BGM for modeling more complex datasets.  Overall, BGM offers a powerful and broadly applicable approach for uncertainty-aware prediction and has the potential for advancing a wide-range of applications in modern data science.

\section{Acknowledgement}
The authors are grateful to Faming Liang, Xiaotong Shen, and Leying Guan for their constructive feedback.

\section{Funding}
Q.L. was partially supported by NIH grant R00 HG013661. W.H.W. was partially supported by NSF-DMS 2310788. 

\section{Disclosure Statement}
The authors report there are no competing interests to declare.

\bibliography{Bibliography-MM-MC}

\newpage
\setcounter{page}{1}
\bigskip
\begin{center}
\end{center}
\section*{Supplementary Materials}
\begin{description}

\item[Appendix A:] Convergence to stationary points

\begin{lemma} If function $f$ is $L$-smooth, then for any $x$, $\Delta$, we have
$$f(x+\Delta)
\;\ge\;
f(x) + \nabla f(x)^{\top}\Delta
\;-\; \frac{L}{2}\|\Delta\|^{2}.$$
\end{lemma}

Proof: since $f$ is $L$-smooth, its gradient is $L$-Lipschitz:
\begin{equation}
\label{eq:Lipschitz-gradient}
\|\nabla f(u)-\nabla f(v)\|
\;\le\;
L\|u-v\|,
\qquad \forall u,v.
\end{equation}

Define $g(t)=f(x+t\Delta)$ for $t\in[0,1]$.  
By the fundamental theorem of calculus,
\begin{equation}
\label{eq:ftc}
f(x+\Delta)-f(x)
= \int_0^1 \nabla f(x+t\Delta)^{\top}\Delta \, dt .
\end{equation}

Add and subtract $\nabla f(x)$ inside the integral:
\begin{align}
f(x+\Delta)-f(x)
&= \int_0^1 
\bigl[\nabla f(x)
+ (\nabla f(x+t\Delta)-\nabla f(x))\bigr]^{\top}\Delta
\, dt \notag \\
&= \nabla f(x)^{\top}\Delta
+ \int_0^1 (\nabla f(x+t\Delta)-\nabla f(x))^{\top}\Delta
\, dt .
\label{eq:split}
\end{align}

Applying Cauchy--Schwarz and the $L$-smoothness inequality
\eqref{eq:Lipschitz-gradient},
\[
\|\nabla f(x+t\Delta)-\nabla f(x)\|
\le L t \|\Delta\|,
\]
we obtain
\begin{equation}
\label{eq:bound-integrand}
(\nabla f(x+t\Delta)-\nabla f(x))^{\top}\Delta
\ge - \|\nabla f(x+t\Delta)-\nabla f(x)\|\,\|\Delta\|
\ge - L t \|\Delta\|^2.
\end{equation}

Substituting \eqref{eq:bound-integrand} into \eqref{eq:split},
\[
f(x+\Delta)-f(x)
\ge \nabla f(x)^{\top}\Delta
\;-\;
\int_0^1 L t \|\Delta\|^2 \, dt .
\]

Since $\int_0^1 L t\,dt = L/2$, we conclude
\[
f(x+\Delta)
\;\ge\;
f(x) + \nabla f(x)^{\top}\Delta
\;-\;
\frac{L}{2}\|\Delta\|^2.
\]

Let $\Delta w_t = \bigl(\eta_t^{(Z)} g_t^{(Z)},\; \eta_t^{(\phi)} g_t^{(\phi)}\bigr).$ $f=\mathcal{J}$, $x=w_t$, and $\Delta=\Delta w_t$. Lemma~1 gives \[
J(w_{t+1})
\;\ge\;
J(w_t)
\;+\;
\nabla J(w_t)^\top \Delta w_t
\;-\;
\frac{L}{2}\,\|\Delta w_t\|^2.
\]
Taking conditional expectation given $\mathcal{F}_t$, \begin{align*}
\mathbb{E}\!\left[J(w_{t+1}) \mid \mathcal{F}_t\right]
\;&\ge\;
J(w_t)
\\
&\quad
+\;
\eta_t^{(Z)}
\left\langle
\nabla_Z J(w_t),
\mathbb{E}\!\left[g_t^{(Z)} \mid \mathcal{F}_t\right]
\right\rangle
\\
&\quad
+\;
\eta_t^{(\phi)}
\left\langle
\nabla_\phi J(w_t),
\mathbb{E}\!\left[g_t^{(\phi)} \mid \mathcal{F}_t\right]
\right\rangle
\\
&\quad
-\;
\frac{L}{2}
\Bigl[
(\eta_t^{(Z)})^2
\mathbb{E}\!\left[\|g_t^{(Z)}\|^2 \mid \mathcal{F}_t\right]
+
(\eta_t^{(\phi)})^2
\mathbb{E}\!\left[\|g_t^{(\phi)}\|^2 \mid \mathcal{F}_t\right]
\Bigr].
\end{align*}.

By using unbiasedness from Assumption~2, the inner products reduce to square norms and we obtain
\begin{align}
\label{eqn:one-step-inequality}
\mathbb{E}\!\left[\mathcal{J}(w_{t+1}) \mid \mathcal{F}_t\right]
&\ge
\mathcal{J}(w_t)
+ \eta_t^{(Z)} \|\nabla_Z \mathcal{J}(w_t)\|^{2}
+ \eta_t^{(\phi)} \|\nabla_\phi \mathcal{J}(w_t)\|^{2}
\notag \\
&\qquad\qquad
- \frac{L}{2}\left(
(\eta_t^{(Z)})^{2} \, \mathbb{E}\|g_t^{(Z)}\|^{2}
+ (\eta_t^{(\phi)})^{2} \, \mathbb{E}\|g_t^{(\phi)}\|^{2}
\right).
\end{align}

Taking total expectations, summing over $t$, and using the bounded variance from Assumption (2) gives
\begin{align}
\sum_{t \ge 0}
\mathbb{E}\!\left[
  \eta_t^{(Z)} \|\nabla_Z \mathcal{J}(w_t)\|^{2}
  + \eta_t^{(\phi)} \|\nabla_\phi \mathcal{J}(w_t)\|^{2}
\right]
&\le
\mathcal{J}^{\star} - \mathcal{J}(w_0)
+ \frac{L}{2}
  \sum_{t \ge 0}
  \left( (\eta_t^{(Z)})^{2} \sigma_Z^{2}
       + (\eta_t^{(\phi)})^{2} \sigma_\phi^{2}
  \right),
\end{align}
where $\mathcal{J}^{\star} = \sup_{w} \mathcal{J}(w) < \infty$ by Assumption (1).

By Assumption (3), $\sum_t \eta_t^{(\cdot)} = \infty$ but $\sum_t (\eta_t^{(\cdot)})^{2} < \infty$;
therefore the right-hand side is finite, which forces
\[
\liminf_{t \to \infty} \mathbb{E}\|\nabla \mathcal{J}(w_t)\|^{2} = 0 .
\]

A standard Robbins--Siegmund supermartingale argument then implies that
$\mathcal{J}(w_t)$ converges almost surely, and
\[
\lim_{t \to \infty} \|\nabla \mathcal{J}(w_t)\| = 0
\quad \text{almost surely}.
\]

\item[Appendix B:] Finite-time rate

Define the block–step-size weights
$\alpha_t := \eta_t^{(Z)} + \eta_t^{(\phi)}$. Starting from the one-step expected-ascent inequality (\ref{eqn:one-step-inequality}), we take total expectations and sum $t=0,\dots,T-1$, then we have
\begin{align}
\sum_{t=0}^{T-1}
\Big( 
\eta_t^{(Z)} \, \mathbb{E}\|\nabla_Z \mathcal{J}(w_t)\|^2
+ \eta_t^{(\phi)} \, \mathbb{E}\|\nabla_\phi \mathcal{J}(w_t)\|^2
\Big)
&\le
\mathbb{E}[\mathcal{J}(w_T)] - \mathbb{E}[\mathcal{J}(w_0)] \nonumber \\
&\quad + \frac{L}{2}
\sum_{t=0}^{T-1}
\Big( 
(\eta_t^{(Z)})^2 \sigma_Z^2
+ (\eta_t^{(\phi)})^2 \sigma_\phi^2
\Big).
\tag{A.17}
\end{align}

Denote ${\eta}_t := \min\{\eta_t^{(Z)},\,\eta_t^{(\phi)}\}$. For any vectors $u$ and $v$, we have
\[
\eta_t^{(Z)} \|u\|^2 
+ \eta_t^{(\phi)} \|v\|^2
\;\ge\;
{\eta}_t \, (\|u\|^2 + \|v\|^2)=\|\nabla \mathcal{J}(w_t)\|^2.
\]
We then obtain
\begin{equation}
\sum_{t=0}^{T-1} 
\eta_t \, \mathbb{E}\|\nabla \mathcal{J}(w_t)\|^{2}
\;\le\;
\mathbb{E}[\mathcal{J}(w_T)] - \mathcal{J}(w_0)
+ \frac{L}{2}
\sum_{t=0}^{T-1}
\left( (\eta_t^{(Z)})^{2}\sigma_Z^{2}
     + (\eta_t^{(\phi)})^{2}\sigma_\phi^{2}
\right).
\tag{A.18}
\end{equation}

Let $R$ be a random index supported on $\{0,1,\ldots,T-1\}$ with
$\mathbb{P}(R = t) = 
\eta_t/{\sum_{t=0}^{T-1}\eta_t} $.

Divide both sides of (A.18) by $\sum_{t=0}^{T-1}\eta_t$:
\begin{align}
\mathbb{E}\|\nabla \mathcal{J}(w_R)\|^{2}&=\frac{
\sum_{t=0}^{T-1} \eta_t \,\mathbb{E}\|\nabla \mathcal{J}(w_t)\|^{2}
}{
\sum_{t=0}^{T-1}\eta_t
}\\
&\le
\frac{
\mathbb{E}[\mathcal{J}(w_T)] - \mathcal{J}(w_0)
}{
\sum_{t=0}^{T-1}\eta_t
}
+
\frac{L}{2}
\frac{
\sum_{t=0}^{T-1}
\left( (\eta_t^{(Z)})^{2}\sigma_Z^{2}
     + (\eta_t^{(\phi)})^{2}\sigma_\phi^{2}
\right)
}{
\sum_{t=0}^{T-1}\eta_t
}.
\tag{2}
\end{align}

Since $\mathbb{E}[\mathcal{J}(w_T)] \le \mathcal{J}^\star$ (compactness of the iterate set),
we arrive at the general finite-time bound:
\[
\mathbb{E}\|\nabla\mathcal{J}(w_R)\|^2
\;\le\;
\frac{
\mathcal{J}^\star - \mathcal{J}(w_0)
}{
\sum_{t=0}^{T-1}\eta_t
}
\;+\;
\frac{
L \sum_{t=0}^{T-1}
\big(
(\eta_t^{(Z)})^2 \sigma_Z^2
+ (\eta_t^{(\phi)})^2 \sigma_\phi^2
\big)
}{2
\sum_{t=0}^{T-1}\eta_t
}.
\tag{A.20}
\]
\item[Appendix C:] Law–level Consistency

We first show that $m(x;\phi)$ is well-defined and uniformly bounded
on each sieve $\Phi_N$.
\begin{lemma}[Coercivity and existence of maximizers in $z$]
Fix $N$ and $\phi \in \Phi_N$. Assume a variance floor
$\sigma_j^2(z;\theta) \ge \sigma^2 > 0$ for all $z,\theta,j$.
With
\[
\log \pi_Z(z) = -\tfrac{1}{2}\|z\|^2 + C,
\]
define
\[
\ell(x,z;\phi)
:= \mathbb{E}_{q_\phi}\!\left[\log P(x \mid z;\theta)\right]
   + \log \pi_Z(z)
   \xrightarrow[\|z\|\to\infty]{} -\infty.
\]
Hence
\[
m(x;\phi) = \sup_z \ell(x,z;\phi) < \infty,
\]
and the supremum is attained (the argmax set is nonempty and compact).
\end{lemma}

Proof: with the variance floor, 
\[
\log p(x \mid z;\theta) \le C_N
\]
uniformly in $z$. Adding $-\tfrac{1}{2}\|z\|^2$ makes
$\ell(x,z;\phi) \to -\infty$ as $\|z\|\to\infty$.
Hence, by the Weierstrass theorem, the supremum is attained. 

\begin{lemma}[Envelope and measurability]
For each fixed $N$, there exists $C_N < \infty$ such that
\[
|m(x;\phi)| \le C_N \quad \text{for all } x \text{ and all } \phi \in \Phi_N.
\]
Moreover, $x \mapsto m(x;\phi)$ is measurable for each $\phi \in \Phi_N$.
\end{lemma}
The bound follows:
\[
\sup_{z}\,\ell(x,z;\phi)
\;\le\;
\sup_{z}\left\{-\tfrac{1}{2}\|z\|^{2}\right\}
\;+\;
C_N
\;=\;
C_N.
\]
Measurability follows because $(x,z,\phi)\mapsto \ell(x,z;\phi)$ is measurable in $x$
and continuous in $(z,\phi)$ on the compact $\Phi_N$; then the supremum over $z$
of a Carathéodory function is measurable.

By Lemma the class $\{\,m(\cdot;\phi) : \phi \in \Phi_N\,\}$ has a
\emph{uniform integrable envelope} and is measurable; thus it is
Glivenko--Cantelli. The KL term is deterministic in $\phi$ and continuous on
compact $\Phi_N$. Based on uniform law of large numbers on $\Phi_N$, 
For each fixed sieve level $N$,
\[
\omega_N
=
\sup_{\phi \in \Phi_N}
\bigl|\widetilde{\mathcal{J}}_N(\phi) - \widetilde{\mathcal{J}}(\phi)\bigr|
\;\xrightarrow{p}\; 0.
\]

Define the \emph{population suboptimality} at the estimator
\[
\Delta_N := \sup_{\phi \in \Phi} \widetilde{\mathcal{J}}(\phi)
- \widetilde{\mathcal{J}}(\widehat{\phi}_N) \;\ge 0.
\]

\[
\Delta_N
= 
\underbrace{\left(\sup_{\phi \in \Phi}\widetilde{\mathcal{J}}
-\sup_{\phi \in \Phi_N}\widetilde{\mathcal{J}}\right)}_{r_N}
\;+\;
\underbrace{\left(\sup_{\phi \in \Phi_N}\widetilde{\mathcal{J}}
- \widetilde{\mathcal{J}}(\widehat{\phi}_N)\right)}_{T_2}.
\]

For any $\phi \in \Phi_N$,
\[
\widetilde{\mathcal{J}}(\phi)
\;\le\;
\widetilde{\mathcal{J}}_N(\phi) + \omega_N,
\qquad
\widetilde{\mathcal{J}}(\widehat{\phi}_N)
\;\ge\;
\widetilde{\mathcal{J}}_N(\widehat{\phi}_N) - \omega_N.
\]

Hence
\[
T_2
\;\le\;
\bigl(\sup_{\phi \in \Phi_N}\widetilde{\mathcal{J}}_N + \omega_N\bigr)
-
\bigl(\widetilde{\mathcal{J}}_N(\widehat{\phi}_N) - \omega_N\bigr)
=
\delta_N^{\mathrm{alg}} + 2\omega_N.
\]

Combine the above inequalities, we have

\[
\Delta_N \;\le\; r_N + 2\omega_N + \delta_N^{\mathrm{alg}}.
\qquad
\]

The oracle inequality bounds the value gap $\Delta_N$.We now show that this forces law‑level convergence through the separation margin $\Delta(\epsilon)$

Let's say the estimator $\hat{\phi}_N$ is $\epsilon$-far from the true model, which is represented as
\[
d(P_{\hat{\phi}_N}, P^\star) \ge \varepsilon.
\]

By the definition of $\Delta(\varepsilon)$, any parameter $\phi$ whose model
is $\varepsilon$--far from the truth must have a population objective value
that is at least $\Delta(\varepsilon)$ below the optimal value.  Therefore,
\[
\widetilde{\mathcal{J}}(\hat{\phi}_N)
\;\le\;
\sup_{\phi\in\Phi}\widetilde{\mathcal{J}}(\phi)
\;-\;
\Delta(\varepsilon).
\]

Rearranging gives
\[
\sup_{\phi\in\Phi}\widetilde{\mathcal{J}}(\phi)
-
\widetilde{\mathcal{J}}(\hat{\phi}_N)
\;\ge\;
\Delta(\varepsilon).
\]

The left-hand side is exactly the definition of $\Delta_N$.  
Hence,
\[
\Delta_N \;\ge\; \Delta(\varepsilon).
\]
Fix $\varepsilon > 0$. Suppose, toward a contradiction, there exists a
subsequence along which
\[
\mathbb{P}\!\left(d(P_{\hat{\phi}_N}, P^\star) \ge \varepsilon\right) \not\to 0.
\]
Since $\Delta_N \;\ge\; \Delta(\varepsilon)$ with non-vanishing probability. But with assumptions, we have
\[
\Delta_N \;\le\; r_N + 2\omega_N + \delta_N^{\mathrm{alg}}
\;\xrightarrow{p}\; 0,
\]
a contradiction since $\Delta(\varepsilon) > 0$ is fixed. Hence for every
$\varepsilon > 0$,
\[
\mathbb{P}\!\left(d(P_{\hat{\phi}_N}, P^\star) \ge \varepsilon\right) \to 0.
\]
This is precisely
\[
d(P_{\hat{\phi}_N}, P^\star) \xrightarrow{p} 0.
\]

\item[Appendix D:] Conditional Excess Risk Bound

We first show the Lipschitz property of $\ell_{KS}$, which is the squared MMD between a Dirac distribution at $y$ and the predictive distribution $r$ using kernel $k$, and we specialize constants to the RBF kernel. For a bounded kernel $k : \mathcal{X}_{\mathcal{B}} \times \mathcal{X}_{\mathcal{B}} \to \mathbb{R}$
and a predictive law $r \in \mathcal{P}(\mathcal{X}_{\mathcal{B}})$,
\[
\ell_{KS}(y,r)
:= k(y,y)
 - 2\,\mathbb{E}_{Y' \sim r}\!\left[k(y,Y')\right]
 + \mathbb{E}_{Y',\,Y'' \sim r}\!\left[k(Y',Y'')\right].
\]

This equals $\operatorname{MMD}^2_k(\delta_y, r)$
 where $\delta_y$ is the Dirac measure at $y$. Thus if $\|k\|_\infty \le K$
(standard MMD bound). Then it is bounded as $\ell_{KS}(y,r) \in [0, 4K]$.

Assume $k$ is bounded by $K$ and Lipschitz in each argument with constant $L_k$
(w.r.t.\ the Euclidean norm on $\mathcal{X}_{\mathcal{B}}$). For $r,s \in \mathcal{P}(\mathcal{X}_{\mathcal{B}})$,
\begin{align}
\bigl|\ell_{KS}(y,r) - \ell_{KS}(y,s)\bigr|
&\le
2\bigl|\mathbb{E}_{Y'\sim r}[k(y,Y')] - \mathbb{E}_{Y'\sim s}[k(y,Y')]\bigr|
\;+\;
\bigl|\mathbb{E}_{r\times r}[k] - \mathbb{E}_{s\times s}[k]\bigr|,
\end{align}
where 
$\mathbb{E}_{r\times r}[k]= \mathbb{E}_{Y',Y'' \sim r}\bigl[\,k(Y',Y'')\,\bigr]$, and 
$\mathbb{E}_{s\times s}[k]=
\mathbb{E}_{Y',Y'' \sim s}\bigl[\,k(Y',Y'')\,\bigr].$

For the first term, $f(\cdot) := k(y,\cdot)$ has $\|f\|_\infty \le K$ and $\mathrm{Lip}(f) \le L_k$. Denote $M=\max\{K,L_k\}$, according to the definition of bounded–Lipschitz distance:
\[
d(r,s)=\sup_{\substack{\|f/M\|_\infty \le 1,\mathrm{Lip}(f/M)\le 1}}
\bigl| \mathbb{E}_r\frac{f}{M} - \mathbb{E}_s\frac{f}{M} \bigr|. \]
Then we have $\bigl|\mathbb{E}_r f - \mathbb{E}_s f\bigr|
\le Md(r,s),$

For the second term, use the triangle inequality:
\[
\bigl|\mathbb{E}_{r\times r}k - \mathbb{E}_{s\times s}k\bigr|
\le 
\bigl|\mathbb{E}_{r\times r}k - \mathbb{E}_{s\times r}k\bigr|
+
\bigl|\mathbb{E}_{s\times r}k - \mathbb{E}_{s\times s}k\bigr|.
\]

Similarly, by holding $s$ or $r$ fixed, we have $\bigl|\mathbb{E}_{r\times r}k - \mathbb{E}_{s\times r}k\bigr|\le Md(r,s)$ and $\bigl|\mathbb{E}_{s\times r}k - \mathbb{E}_{s\times s}k\bigr|\le Md(r,s)$, combining     together, we have $\bigl|\mathbb{E}_{r\times r}k - \mathbb{E}_{s\times s}k\bigr|
\le 2Md(r,s)$.

Combining the above two terms together, we have
\begin{equation}
\label{eqn:lip_inequ}
\bigl|\ell_{KS}(y,r) - \ell(y,s)_{KS}\bigr|
\;\le\;
4M\, d(r,s),
\qquad
M := \max\{K, L_k\}.
\end{equation}

Fix $x_{\mathcal{A}} \in \mathcal{X}_{\mathcal{A}}^{\circ}$ and $y \in \mathcal{X}_{\mathcal{B}}$.
By the Lipschitz property in (\ref{eqn:lip_inequ}),
\[
\bigl| \ell_{KS}(y, g_{p_{\hat{\phi}_N}}(x_{\mathcal{A}}))
     - \ell_{KS}(y, g_{p^\star}( x_{\mathcal{A}})) \bigr|
\;\le\;
L_{\ell}\, d\!\bigl(g_{p_{\hat{\phi}_N}}(x_{\mathcal{A}}), g_{p^\star}(x_{\mathcal{A}})\bigr)
\;\le\;
L_{\ell}\, \varepsilon_N^{\mathrm{cond}}.
\]

Taking expectation with respect to 
$(X_{\mathcal{A}}, X_{\mathcal{B}}) \sim P^{\star,\circ}$,
\[
\bigl| \mathcal{R}_{P^\star}^{\circ}(g_{p_{\hat{\phi}_N}})
      - \mathcal{R}_{P^\star}^{\circ}(g_{p^\star}) \bigr|
=
\bigl| 
\mathbb{E}_{P^\star,\circ}[
  \ell_{KS}(X_{\mathcal{B}}, g_{p_{\hat{\phi}_N}}(X_{\mathcal{A}}))
 -\ell_{KS}(X_{\mathcal{B}}, g_{p^\star}(X_{\mathcal{A}}))
] \bigr|
\;\le\;
\mathbb{E}_{P^\star,\circ} \bigl[L_{\ell}\, \varepsilon_N^{\mathrm{cond}}\bigr]
=
L_{\ell}\, \varepsilon_N^{\mathrm{cond}}.
\]

So we finally have
\[
\operatorname{Excess}_{P^\star}^{\circ}(g_{p_{\hat{\phi}_N}})
\;\le\;
L_{\ell}\, \varepsilon_N^{\mathrm{cond}}.
\]

\item[Appendix E:] Baseline Methods

This appendix provides implementation details for all baseline methods used in the empirical evaluation, including point prediction methods, conformal prediction methods for interval estimation, and data imputation baselines. All baselines were implemented using standard and publicly available libraries.

The Linear Regression was implemented using the \texttt{LinearRegression} class from the \texttt{scikit-learn} library~\citep{pedregosa2011scikit} with default parameters. We employed random forest regression as a flexible ensemble-based nonparametric baseline \citep{breiman2001random}, implemented using the \texttt{RandomForestRegressor} class from the \texttt{scikit-learn} library with default hyperparameters.  Gradient boosted decision trees were implemented using the XGBoost Python package \citep{chen2016xgboost}. We trained an \texttt{XGBRegressor} with squared error loss, using 500 boosting iterations, a learning rate of 0.05, and a maximum tree depth of 4. Subsampling was applied to both observations and features, with subsample and column subsample ratios set to 0.8. These settings follow common practice to balance predictive accuracy and regularization. XGBoost represents a strong ensemble-based machine learning baseline that sequentially improves predictions by fitting trees to residual errors.


All conformal prediction methods were implemented using the official github repository (\url{https://github.com/LeyingGuan/LCPexperiments}) for localized conformal prediction (LCP)~\citep{guan2023localized}. The vanilla CP constructs prediction intervals based on absolute residuals from a fitted point predictor using a held-out calibration set. The nonconformity score is defined as the absolute prediction error. LW-CP normalizes residuals by an estimated local noise level to account for heteroscedasticity \citep{lei2018distribution}. Local variance estimates are obtained via nearest-neighbor smoothing in the predictor space. QR-CP constructs prediction intervals by conformalizing estimated conditional quantiles \citep{romano2019conformalized}. Lower and upper quantile regressions were trained using the neural network predictor, and conformal scores were formed based on quantile violations. LWQR-CP further adjusts quantile-based conformal scores using local variability estimates, improving adaptivity in heterogeneous settings. Localized CP \citep{guan2023localized} extends standard CP by weighting calibration samples according to their similarity to the test point. Similarity was measured using Euclidean distance in the learned feature representation of the neural network predictor. Applying localization to the above score constructions yields three additional baselines: LW-LCP, QR-LCP, and LWQR-LCP. These methods combine the respective nonconformity scores with localized weighting schemes to improve empirical adaptivity.

Missing values from MNIST testing set were imputed using the empirical mean of each variable computed from the observed entries. This mean imputation baseline provides a simple and fast reference method and was implemented using \texttt{SimpleImputer} class from  the \texttt{scikit-learn} library. MICE~\citep{van2011mice} method teratively imputes missing values using conditional regression models for each variable. We implemented MICE using \texttt{IterativeImputer} class from  the \texttt{scikit-learn} library with default settings.

\end{description}

\clearpage
\appendix

\setcounter{table}{0}
\renewcommand{\thetable}{S\arabic{table}}

\setcounter{figure}{0}
\renewcommand{\thefigure}{S\arabic{figure}}

\section*{Supplementary Tables}

\begin{table}[!htbp]
\centering
\begin{threeparttable}
\caption{Comparison of interval estimation performance across BGM and VAEAC. Metrics include Pearson correlation (PCC), Spearman correlation (SCC), empirical marginal coverage, and average prediction interval length (ave.PI) at different dimensions $p$. The experimental setting is the same as Table~\ref{tab:interval_estimate_comparison}. VAEAC failed to capture meaningful uncertainty based on the multiple imputations.}
\label{tab:vaeac}

\begin{tabular}{l cc cc cc}
\toprule
& \multicolumn{2}{c}{$p=50$} & \multicolumn{2}{c}{$p=100$} & \multicolumn{2}{c}{$p=300$} \\
\cmidrule(lr){2-3}\cmidrule(lr){4-5}\cmidrule(lr){6-7}
Metric & VAEAC & BGM & VAEAC & BGM & VAEAC & BGM \\
\midrule
PCC$\uparrow$      & 0.015  & \textbf{0.937} & 0.219 & \textbf{0.874} & -0.100 & \textbf{0.863} \\
SCC$\uparrow$      & 0.030  & \textbf{0.987} & 0.175 & \textbf{0.935} & -0.014 & \textbf{0.941} \\
Coverage           & 0.050  & \textbf{0.944} & 0.056  & \textbf{0.950} & 0.024  & \textbf{0.966} \\
ave.PI & 0.043  & \textbf{1.450} & 0.060  & \textbf{1.576} & 0.038  & \textbf{1.694} \\
\bottomrule
\end{tabular}

\begin{tablenotes}[para,flushleft]
\footnotesize
\textit{Note}: $^{\ast}$ Coverage values closest to the nominal level 0.95 are highlighted in bold. Average interval lengths closest to the ground truth are highlighted in bold.
\end{tablenotes}
\end{threeparttable}
\end{table}

\clearpage

\section*{Supplementary Figures}

\begin{figure}[!htbp]
\centering
\includegraphics[width=6in]{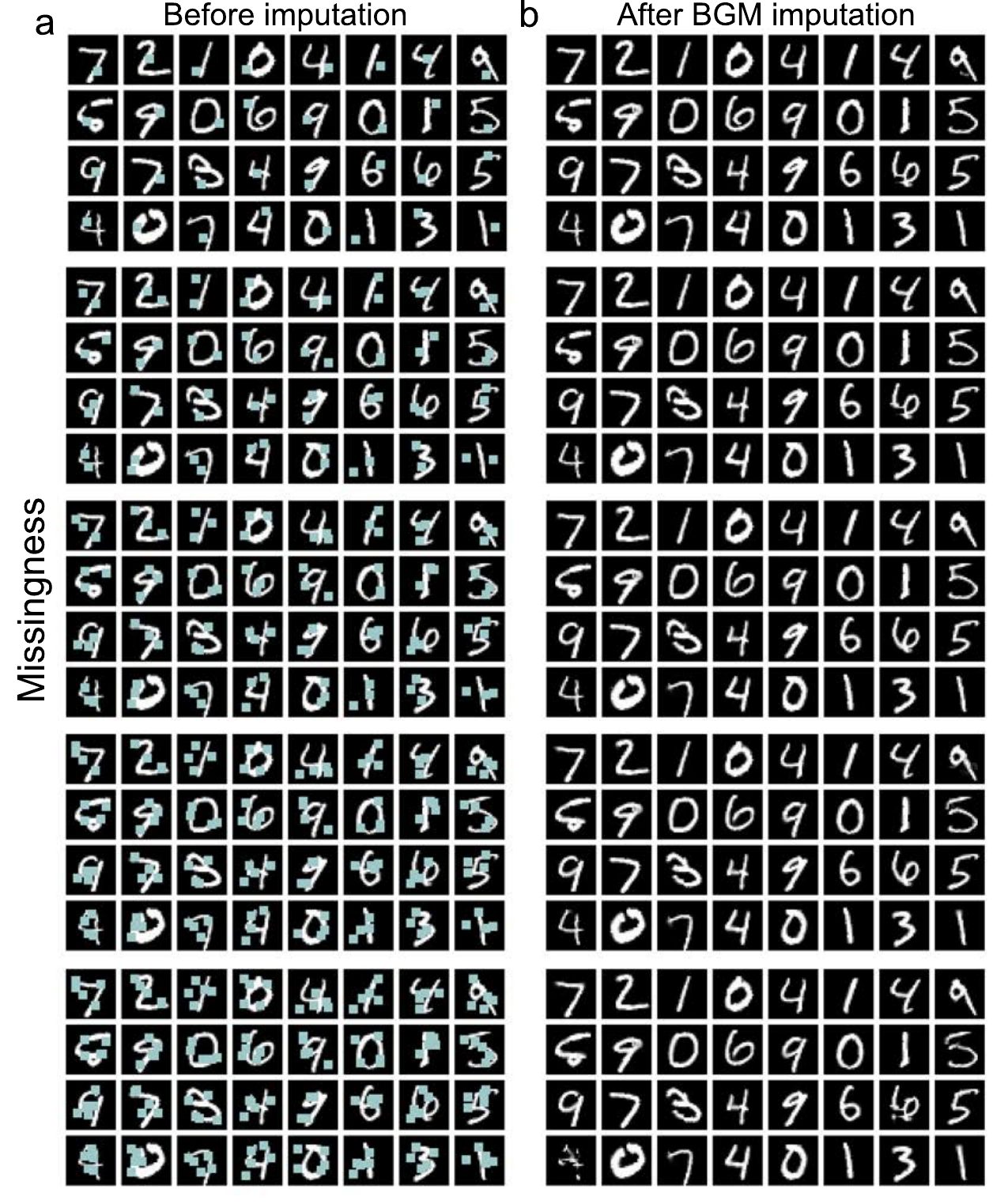}
\caption{More imputation results of BGM on MNIST testing set with different level of missingness. (a) Test images with different level of missingness. (b) Imputed images with a trained BGM model.  \label{fig:Missingness}}
\end{figure}

\begin{figure}[!htbp]
\centering
\includegraphics[width=6in]{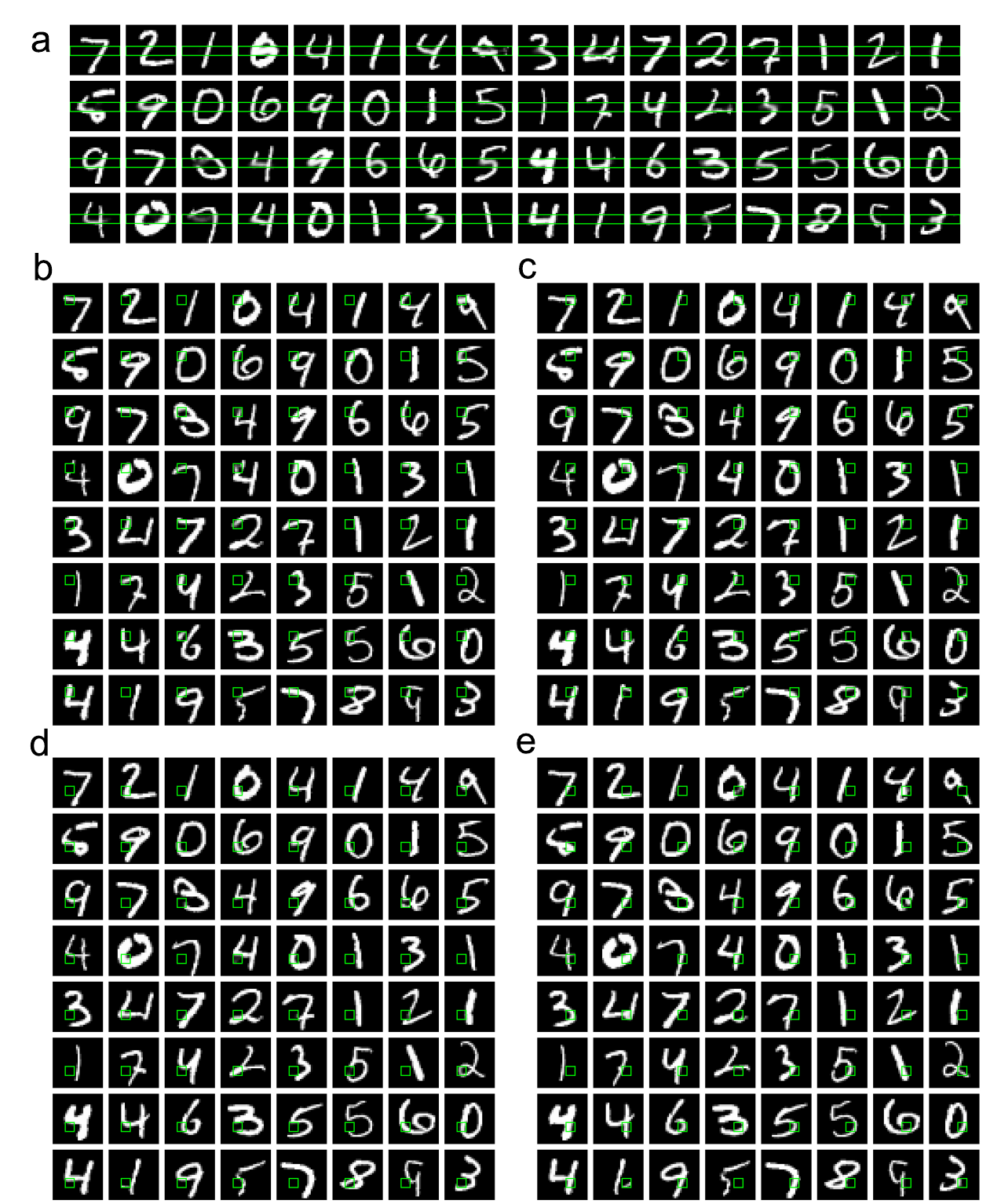}
\caption{More imputation results of BGM on MNIST testing set with different missingness patterns. (a) A $5 \times 13$ stripe mask in the middle. (b) A $5 \times 5$ square mask in the upper left. (c) A $5 \times 5$ square mask in the upper right (d) A $5 \times 5$ square mask in the lower left. (e) A $5 \times 5$ square mask in the lower right. Note that the pixels within the green rectangles are imputed by BGM posterior mean. \label{fig:More_imp}}
\end{figure}



\end{document}